\def\WAMExtendedVersion{1}
\def\WAMArxivRelease{1}
\documentclass[letterpaper,10pt,conference]{ieeeconf}

\IEEEoverridecommandlockouts
\usepackage[T1]{fontenc}
\usepackage{amsmath}
\usepackage{amssymb}
\usepackage{balance}
\usepackage{booktabs}
\usepackage{cite}
\usepackage{colortbl}
\usepackage{graphicx}
\usepackage{longtable}
\usepackage[protrusion=false]{microtype}
\usepackage{placeins}
\usepackage{xcolor}
\ifdefined\ICRASubmissionOnly
  \usepackage[draft,bookmarks=false]{hyperref}
\else
  \usepackage[breaklinks,colorlinks,citecolor=blue,linkcolor=blue,urlcolor=blue]{hyperref}
  \def\WAMExtendedVersion{1}
\fi
\usepackage[capitalize]{cleveref}

\providecommand{\citep}[1]{\cite{#1}}

\crefname{figure}{Fig.}{Figs.}
\crefname{table}{Tab.}{Tabs.}
\Crefname{figure}{Fig.}{Figs.}
\Crefname{table}{Tab.}{Tabs.}
\crefname{appendix}{Appendix}{Appendices}
\Crefname{appendix}{Appendix}{Appendices}

\definecolor{tablegray}{gray}{0.94}
\definecolor{metricgreen}{RGB}{0,128,0}
\definecolor{metricred}{RGB}{180,0,0}


\renewcommand{\keywords}[1]{}

\title{\LARGE \bf
Beyond Visual Quality: A Study of\\
Test-Time Planning with World Action Models}

\ifdefined\WAMArxivRelease
\author{Jianhao Yuan$^{1}$, Yu Yuan$^{2}$, Benjamin Ramtoula$^{1}$, Lukas Vierling$^{1}$,\\
Paul Newman$^{1}$, Lars Kunze$^{3}$, Philip Torr$^{1}$, Daniele De Martini$^{1}$\\[4pt]
{\small $^{1}$University of Oxford\qquad $^{2}$Purdue University\qquad $^{3}$UWE Bristol}}
\hypersetup{pdftitle={Beyond Visual Quality: A Study of Test-Time Planning with World Action Models},
pdfauthor={Jianhao Yuan, Yu Yuan, Benjamin Ramtoula, Lukas Vierling, Paul Newman, Lars Kunze, Philip Torr, Daniele De Martini}}
\else
  \author{Anonymous Authors}
\fi

\begin{document}
\maketitle
\ifdefined\WAMArxivRelease
  \thispagestyle{plain}
  \pagestyle{plain}
\else
  \thispagestyle{empty}
  \pagestyle{empty}
\fi

\bstctlcite{IEEEexample:BSTcontrol}

\begin{abstract}
World action models generate actions together with visual predictions of their
consequences. These paired outputs create the potential for planning by
sampling multiple actions from one state, comparing their imagined outcomes, and
choosing the action with the most promising predicted outcome. However, how
to use imagined futures to guide action selection remains unclear. We examine
this planning potential empirically. First, we estimate an oracle upper bound
on selection by choosing the sampled candidate whose realised outcome is best.
In a controlled same-state analysis, this choice raises success from $68.9\%$
under uniform random selection to $79.2\%$. We then test selectors based on visual quality,
physical consistency, and task progression as controlled interventions.
Some tested selectors yield higher observed success, but the gains are uneven and the
matched selectors leave much of the measured opportunity unrecovered. To
investigate this gap, we examine whether sampled actions lead to different
outcomes, whether these differences are visible in the predictions, and
whether a score recognises them. Counterfactual branching from the same states shows that selection
opportunity is concentrated in relatively few decisions in the initial candidate sets.
Action spread and outcome coverage need not increase together. In a further
evaluation across trajectory phases with complete action execution, the tested
scores again recover little of the available improvement despite a small gain
from learned value.
These findings distinguish producing consequential action choices from
recognising them in generated futures, motivating the evaluation of WAM
predictions through their usefulness for decisions rather than visual quality
alone.
\end{abstract}

\section{Introduction}
\label{sec:intro}

World action models (WAMs) jointly produce actions and visual predictions of
their consequences~\citep{dreamzero2024,cosmospolicy2026,cen2025worldvla,li2026lingbotva}.
Pairing actions with visual futures creates the potential for planning through
action selection. An agent can sample actions from one state, compare their
imagined outcomes, and execute the one associated with the most promising
future, steering behaviour at test time without retraining the WAM.

\begin{figure}[t]
\centering
\includegraphics[width=\columnwidth]{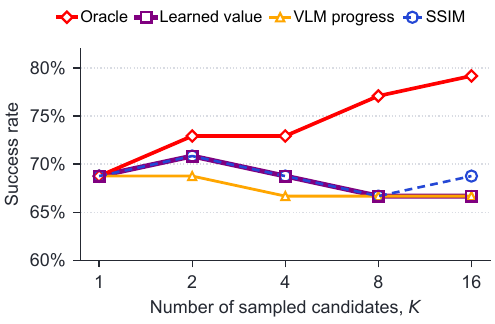}
\caption{\textbf{Substantial potential remains in action planning.}
  Searching among sampled actions with access to their ground truth outcomes
  indicates substantial potential to improve task success. However, the tested
  learned value estimates, visual scores and VLM judgements leave much of this
  potential unrealised as the number of candidates increases. This offline
  comparison ranks the same candidate sets at identical states under a common
  continuation policy. Coincident curves have equal success.}
\label{fig:trajectory_coverage}
\end{figure}

However, how to realise this planning potential through generated futures
remains an open question. To quantify the potential gain from selection, we
first execute sampled actions from the same state under a common
continuation policy. Their realised outcomes define an oracle upper bound on
selection. For this comparison, each decision state has a precomputed set of
paired future and action samples that is shared by every selector.
In our primary study of Cosmos Policy~\citep{cosmospolicy2026} in
RoboCasa~\citep{robocasa2024}, this oracle raises success from $68.9\%$ under
uniform random selection to $79.2\%$ within these candidate sets. Although
unavailable at deployment, the oracle provides a reference for assessing how
much improvement a practical score can recover.

\begin{figure*}[t]
\centering
\includegraphics[width=0.985\textwidth]{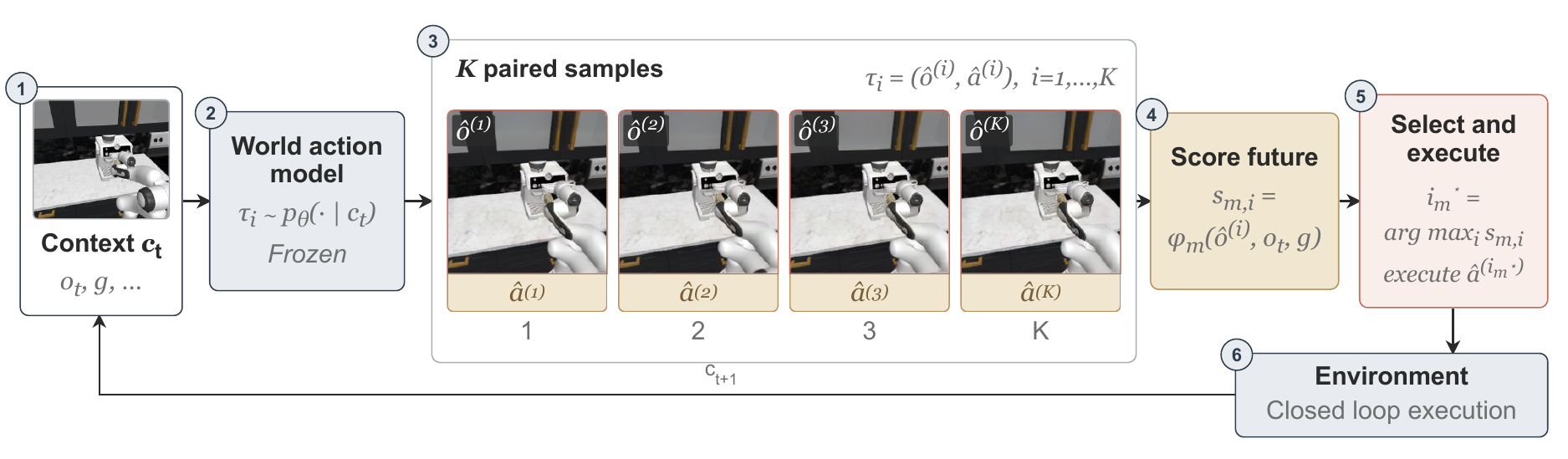}
\caption{\textbf{Best of $K$ selection as a controlled study.}
  Context $c_t$ includes the current observation $o_t$ and task instruction $g$.
  A fixed WAM draws pairs $\tau_i=(\hat{o}^{(i)},\hat{a}^{(i)})$ of a predicted
  visual future $\hat{o}^{(i)}$ and the executable part $\hat{a}^{(i)}$ of its
  associated action sequence. Rule $m$ computes
  $s_{m,i}=\phi_m(\hat{o}^{(i)},o_t,g)$, selects
  $i_m^\star=\arg\max_i s_{m,i}$, executes $\hat{a}^{(i_m^\star)}$, and
  advances the environment to context $c_{t+1}$ for closed loop execution.
  The samples show one recorded RoboCasa decision. Generation and execution
  remain fixed across rules, isolating the choice of score.}
\label{fig:method}
\end{figure*}

We then evaluate whether scores based on visual quality, physical consistency,
task progression, \ifdefined\WAMArxivRelease and \else robot reward and \fi
learned value can recover this improvement
from generated futures.
As illustrated in \cref{fig:method}, standard best of $K$ selection serves as
a controlled testbed, with the pretrained WAM and its sampling
procedure held fixed. Closed loop evaluations measure task success, while
matched comparisons rank the same candidates used to estimate the oracle.
Some scores yield higher observed success, but unevenly across tasks.
As shown in
\cref{fig:trajectory_coverage}, the matched comparisons of learned value, image
similarity and VLM judgements leave much of the available opportunity unrecovered.

We hypothesise that the usefulness of generated futures for action selection
depends on the diversity of action consequences, the visibility of these
consequences in the predictions, and the alignment of the scoring rule with task
outcomes. We refer to these factors as \emph{task consequence diversity},
\emph{outcome observability} and \emph{ranking alignment}, respectively.
Task consequence diversity concerns differences in task outcome among candidate
actions from the same state. Outcome observability concerns the evidence of
these differences in the predicted futures. Ranking alignment concerns whether
a score orders candidates in agreement with their realised outcomes.
We investigate these factors through executed candidate outcomes, which
measure the availability of useful alternatives, and visual diagnostics of
prediction variation and agreement with a recorded observation. Comparing scores
with executed outcomes tests whether they recognise successful alternatives.
Distinguishing an outcome obscured by the prediction from one
overlooked by the evaluator remains an open part of the diagnosis.

The analysis shows that selection opportunity is concentrated in relatively
few decisions in these initial candidate sets. Most decisions contain only successful or
only failed continuations, so changing the ranking cannot improve terminal
success there. Where outcomes differ, successful alternatives are a minority
and the tested scores often miss them. Selecting more widely separated actions
within each candidate set leaves oracle success unchanged relative to
ordered subsets. A further evaluation samples decisions across trajectory
phases and executes complete action chunks. Learned value provides a small
observed gain, but the tested scores still recover little of the available
improvement. Thus, larger distances between sampled action sequences do not
necessarily provide more successful alternatives, and a visual score may fail
to recognise the successful alternatives that are available.
These findings distinguish producing successful actions from recognising them
in generated futures. WAM development can improve both, rather than pursuing
diversity for its own sake. If every candidate succeeds, no reranking is needed.
Our key contributions are as follows.
\begin{itemize}
\setlength{\topsep}{4pt}
\setlength{\itemsep}{0pt}
\setlength{\parsep}{0pt}
  \item We quantify the planning potential available within sampled WAM actions
  using outcome informed upper bounds on selection.
  \item We evaluate existing visual and value scores through online control
  and matched candidate ranking, separating observed gains from unrecovered
  selection opportunity.
  \item We examine how action spread and score rankings relate to successful
  alternatives, and test the selection gap across trajectory phases with
  complete action chunks.
\end{itemize}

\section{Related Work}
\label{sec:related}

\subsection{Predictive Models and World Action Models}
Learned world models support control by planning through latent dynamics, as in
PlaNet~\citep{hafner2019planet} and TD-MPC~\citep{hansen2022tdmpc}, or through
predicted visual observations, as in Deep Visual Foresight
~\citep{finn2017visualforesight} and Visual Foresight
~\citep{ebert2018visualforesight}. UniPi~\citep{du2023unipi} recovers actions from
generated video plans, while UniSim~\citep{yang2024unisim} learns an interactive
simulator for policy training. Contemporary WAMs bring prediction and action
generation into one system. Examples include WorldVLA~\citep{cen2025worldvla},
DreamZero~\citep{dreamzero2024}, Cosmos Policy~\citep{cosmospolicy2026},
LingBot-VA~\citep{li2026lingbotva}, and Motus~\citep{bi2025motus}. Recent work has
made these predictions more closely tied to control. HarmoWAM
~\citep{feng2026harmowam} uses adaptive experts, DynamicWAM
~\citep{lou2026dynamicwam} introduces motion conditioning, SelfWAM
~\citep{pan2026selfwam} adds self grounding, and Qiu et al.
~\citep{qiu2026agra} repurpose representation alignment. Fast-WAM
~\citep{fastwam2026}, ForeWAM~\citep{huang2026forewam}, and RIFT
~\citep{zhang2026rift} further examine whether predictive information can support
a policy without decoding a visual rollout at every control step. World Action
Verifier~\citep{liu2026wav} instead separates state plausibility from action
reachability when verifying a world model. These results concern how predictive
representations are learned, consumed, or verified. Our study addresses a
different question. Given the paired futures and actions already produced by a
fixed WAM, we test whether the futures provide evidence for selecting the better
action.

\subsection{Evaluation of Generated Futures}
Assessing these predictions requires distinguishing visual plausibility from
usefulness for control.
VBench~\citep{huang2024vbench} decomposes generated video quality into
perceptual and temporal dimensions, while VideoScore~\citep{he2024videoscore}
learns multi aspect ratings from human feedback. VideoPhy
~\citep{bansal2024videophy} evaluates physical commonsense in generated videos,
while WMReward~\citep{yuan2026wmreward} uses a latent world model to score
physical plausibility. Robometer~\citep{robometer2026} instead
learns a general purpose reward model from robot trajectory comparisons. A
complementary line of work evaluates functional utility through planning,
action recovery, or action conditioned prediction. This includes the
control-centric benchmark of Tian et al.~\citep{tian2023vp2},
WorldSimBench~\citep{worldsimbench2024}, WorldArena~\citep{worldarena2026},
MiraBench~\citep{mirabench2026}, and RoboWorld~\citep{roboworld2026}.
RoboWM-Bench~\citep{jiang2026robowmbench} further tests whether generated
manipulation behaviours can be converted into executable actions that complete
the intended task. These works expose an important distinction for action
selection. A future can score highly as a video without revealing which paired
action will complete a task. We therefore evaluate representative metric
families as online ranking rules, then compare selected scores on identical
branches with known terminal outcomes. This separates observed control
performance from the opportunity available within a candidate set.

\subsection{Test Time Action Selection and Verification}
Test time selection can improve a fixed policy by generating alternatives and
choosing among them. V-GPS~\citep{vgps2024} reranks actions with a learned value
function, while FOREWARN~\citep{forewarn2025} aligns a VLM with predicted latent
states. CoVer~\citep{kwok2026cover} combines task conditioned verification with
expanded instruction and action proposals. VERITAS~\citep{zhang2026veritas}
compares projected sampled action chunks with a VLM-generated visual trace.
Recent methods for
predictive policies rank candidates using future
consensus~\citep{ruan2026futurecompatible}, cross view
geometry~\citep{zhao2026geobon}, inverse dynamics
consistency~\citep{seo2026acid}, or task progress prediction and consistency
under repeated denoising~\citep{zhou2026tau0wm}.
These works report settings in which grounded verification improves control.
GeoBoN~\citep{zhao2026geobon} also uses fixed candidate sets for offline
diagnostics. Our focus is the diagnostic question behind
such gains. Same state terminal outcomes measure whether a candidate set
contains successful alternatives, and matched scoring tests whether a rule
recognises them. This complements the development of stronger selectors by
separating available opportunity from the gain a selector recovers.

\section{Selection Opportunity from Stochastic Sampling}
\label{sec:setup}

We start by asking whether sampling different actions can improve task success.
The first experiment restarts entire episodes from the same initial scene with
different policy seeds, asking whether another complete run succeeds. The
second tests different candidate actions at one recorded decision state under
a common continuation policy, asking whether one changed choice affects success.

Our primary evaluation uses Cosmos Policy 2B~\citep{cosmospolicy2026} in
RoboCasa~\citep{robocasa2024}. The model predicts 32 actions, executes the first
16, and replans until success or the task step limit. A separate full chunk
evaluation in \cref{sec:full_chunk} executes all 32 actions. The trajectory and closed loop
evaluations cover the same 24 tasks. The same state experiments use the
targeted subsets of tasks and states described below. We reserve \emph{candidate}
for a paired future and action sample at one decision, and \emph{trajectory}
for a complete policy run.

\begin{figure*}[t]
\centering
\includegraphics[width=0.98\textwidth]{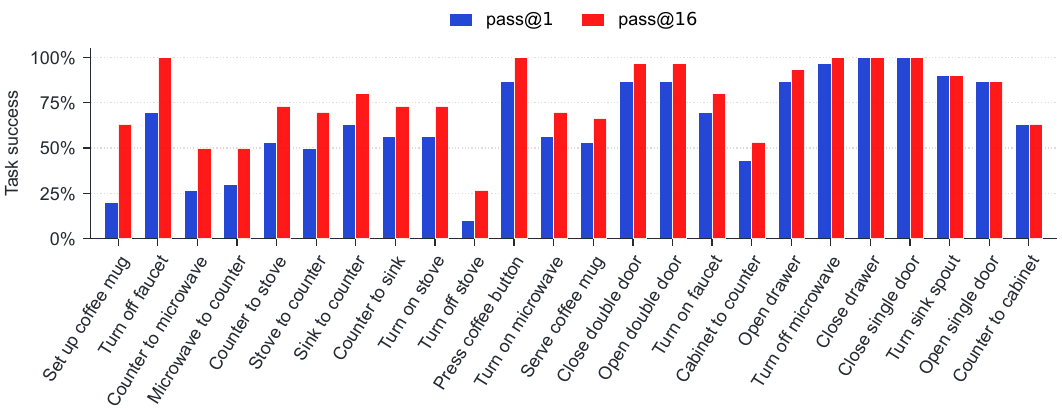}
\caption{\textbf{Trajectory coverage by task.}
  Bars compare pass@$1$ with retrospective pass@$16$. All 24 tasks use the
  same 30 scenes and the same ordered set of 16 seeds. Tasks are ordered by the
  increase in coverage.}
\label{fig:task_coverage}
\end{figure*}

\subsection{Trajectory Level Opportunity}

For each of the 24 tasks, we run 30 fixed scenes under 16 policy noise seeds,
giving 11,520 complete trajectories. Environment identity is held fixed within
each task and scene, and only the policy seed changes. This can change action
choices throughout the episode. We keep the seed order fixed and examine the
first $K=1,2,4,8,16$ runs. Using their recorded terminal outcomes, pass@$K$
counts a scene as covered if at least one of these $K$ trajectories succeeds.
We average over all 720 task and scene pairs, which also equals the task mean
because each task contributes 30 scenes. Pooled pass@$K$ rises from $64.3\%$
at $K=1$ to $68.3\%$, $71.4\%$,
$74.7\%$, and $77.4\%$ at $K=2,4,8,16$.
As shown in \cref{fig:task_coverage}, gains vary across tasks.
Setting up a coffee mug rises from $20.0\%$ to $63.3\%$
and turning off a faucet from $70.0\%$ to $100\%$. Some tasks are already at
ceiling, but ceiling effects do not explain every unchanged result. Moving an
object from the counter to the cabinet remains at $63.3\%$.

\subsection{Same State Opportunity}

To test a single action choice, we return to a recorded decision state rather
than restarting the episode. From its context $c_t$, the fixed WAM is sampled
repeatedly with different seeds to produce paired futures and action sequences.
We restore the same simulator state before executing each candidate's first
16 actions, then run the same continuation policy and seed schedule until
success or the task step limit. We call this completed rollout a candidate
continuation. The policy is the same across candidates, although its subsequent
actions can differ because the initial choice changes the state it observes.

The initial experiment uses 4 tasks and 3 behaviour trajectories per task.
We record the first 4 replanning states encountered on each trajectory, giving
48 decision states. The first candidate drives the behaviour trajectory, so
testing alternatives does not change which later states are recorded. At each
state we use the first 16 candidates in the fixed sampling order, giving 768
continuations. The states, candidate order, continuation policy and scoring
rules were fixed before scores were joined to terminal outcomes.

Restored state hashes and replays of the first candidate's 16 actions passed
collection checks, testing restoration and isolation between executions.
They do not establish repeatability of every complete continuation. A follow-up
reproduces one full continuation four times, including its actions and outcome,
but does not certify every archived continuation.

For each state, we compare the first $K$ candidates. Uniform selection succeeds
with probability equal to the fraction whose continuations succeed. The
outcome oracle succeeds whenever at least one candidate continuation succeeds.
We average these quantities over the same 48 states and call their difference
\emph{local selection headroom}. Unlike pass@$K$ over whole trajectories, this
oracle changes only the initial candidate choice, not later decisions.
A state exhibits \emph{task consequence diversity} when its candidate
continuations include both success and failure. As shown in
\cref{fig:trajectory_coverage}, oracle success reaches
$79.2\%$ at $K=16$, compared with $68.9\%$ under uniform selection.
At the $K=4$ used by the online study, the
corresponding oracle and uniform success rates are $72.9\%$ and $69.3\%$.
Within these recorded states, the fraction with both outcomes grows from
$6.3\%$ at $K=4$ to $12.5\%$ at $K=16$. All 6 such states at $K=16$
occur in the 2 nonsaturated tasks. These results motivate testing whether the generated futures reveal which
candidate actions achieve better outcomes.
\ifdefined\WAMExtendedVersion
Complete protocols appear in \cref{app:opportunity}.
\fi

\section{Evaluating Existing Selection Metrics}
\label{sec:results}

Trajectory coverage and same state oracles show that sampling can
produce useful alternatives. Realising this gain at test time requires a score
that identifies the better candidate from its predicted future. We audit
existing metrics while holding the WAM, sampler, action horizon, and execution
loop fixed. In \cref{tab:closed_loop_by_difficulty}, we compare selection rules
across tasks grouped by the fixed reference success defined below.

\begin{table}[t]
\centering
\caption{\textbf{Closed-loop success by task difficulty.}
  Fixed difficulty groups are defined in \cref{sec:controlled_selection}.
  Bold marks the highest observed success in each column.}
\label{tab:closed_loop_by_difficulty}
\footnotesize
\setlength{\tabcolsep}{2.2pt}
\renewcommand{\arraystretch}{1.0}
\begin{tabular*}{\columnwidth}{@{\extracolsep{\fill}}lrrrr@{}}
\toprule
\textbf{Selection rule} & \multicolumn{4}{c}{\textbf{Task success (\%)}} \\
\cmidrule(l){2-5}
& \textbf{Hard} & \textbf{Medium} & \textbf{Easy} & \textbf{Overall} \\
\midrule
\rowcolor{tablegray}[0pt][0pt]Fixed candidate, $K=1$ & 31.0 & 66.9 & 94.3 & 65.9 \\
\midrule
\rowcolor{tablegray}[0pt][0pt]Learned value & \textbf{33.2} & 67.1 & \textbf{95.3} & \textbf{66.8} \\
Background consistency & 31.0 & 69.3 & 93.1 & 66.7 \\
Laplacian sharpness & 31.7 & 67.6 & 93.1 & 66.1 \\
Gradient sharpness & 30.0 & 67.5 & 94.0 & 65.8 \\
Spatial complexity & 30.7 & 67.6 & 93.1 & 65.8 \\
Aesthetic quality & 31.0 & 66.7 & 93.7 & 65.7 \\
SSIM & 31.3 & 66.4 & 93.4 & 65.5 \\
Subject consistency & 32.3 & 66.5 & 91.7 & 65.3 \\
Image quality & 30.7 & 66.5 & 93.1 & 65.3 \\
Frame difference & 30.0 & 66.2 & 94.0 & 65.3 \\
Foreground centroid jerk & 29.0 & 67.1 & 92.9 & 65.1 \\
Temporal flicker & 30.0 & 66.2 & 92.0 & 64.7 \\
VLM controllability & 30.0 & \textbf{69.5} & 90.0 & 65.6 \\
VLM progress value & 32.3 & 67.6 & 90.9 & 65.6 \\
VQAScore & 28.3 & 68.9 & 90.0 & 64.9 \\
VLM progress & 29.0 & 66.9 & 91.7 & 64.7 \\
VideoPhy & 30.0 & 66.0 & 90.9 & 64.3 \\
\bottomrule
\end{tabular*}
\end{table}

\subsection{Controlled Selection}
\label{sec:controlled_selection}

At decision $t$, context $c_t$ contains the current observation, task
instruction, and other information available to the WAM. Different sampling
seeds produce $K$ paired futures and action sequences. We denote each pair by
$\tau_i=(\hat{o}^{(i)},\hat{a}^{(i)})$, using $\hat{a}^{(i)}$ for the
executed part of the predicted action sequence. Selection rule $m$
evaluates each predicted future and assigns the score
$s_{m,i}=\phi_m(\hat{o}^{(i)},o_t,g)$. It selects the candidate indexed by
\begin{equation}
i_m^\star=\arg\max_i s_{m,i}
\end{equation}
and executes the paired actions $\hat a^{(i_m^\star)}$. The procedure then
repeats from the new observation, as illustrated in \cref{fig:method}.
Appearance scores see only the predicted future. Similarity scores may also
use the current image $o_t$, while task progress scores receive the instruction
$g$. No external selector receives the paired action, simulator state,
realised observation, or eventual outcome. The learned value is the pretrained
return estimate emitted by Cosmos Policy with each pair, not a value function
refitted to our continuation. No selector is fitted to the RoboCasa success
labels.

Each online rule is evaluated on 24 tasks with 50 trials per task, except
learned value, which pools two independently run panels. At every decision,
Cosmos Policy produces $K=4$ paired samples, executes the first 16 actions of
the selected candidate, and replans. Seeds, sampler, horizon, and evaluation code
remain fixed across rules. The $K=1$ reference follows the same initial
condition schedule but is run separately, so it is a single sample reference
rather than a random candidate from the saved $K=4$ sets. Task difficulty is
fixed using this reference success. Hard tasks have success
below $50\%$, medium tasks have success from $50\%$ to below $80\%$, and easy
tasks have success of at least $80\%$. These groups contain 6, 11, and 7 tasks,
respectively, and remain unchanged across selectors.
\ifdefined\WAMExtendedVersion
Additional selector details appear in \cref{app:complete_selector_results}.
\fi

The audit contains nine image or motion scores and five components implemented
from VBench~\citep{huang2024vbench}, together with the learned value. The
released interface exposes one decoded future frame. The \emph{repeated
frame} configuration repeats it when a score expects video.
The \emph{decoded slot} configuration supplies the current
observation followed by the four decoded images associated with the primary
camera future slot. This slot is trained from four copies of
one timestep, not four successive timesteps~\citep{cosmospolicy2026}. We use
these decoded slot images to evaluate VideoPhy
~\citep{bansal2024videophy},
\ifdefined\WAMArxivRelease\else
WMReward~\citep{yuan2026wmreward}, Robometer~\citep{robometer2026},
\fi
and four task conditioned scores implemented with
Qwen3-VL 4B Instruct~\citep{bai2025qwen3vl}. VLM controllability and VLM
progress value parse numerical ratings. VLM progress uses a Yes-answer
probability for improvement relative to the current image, while the Qwen
implementation of VQAScore~\citep{lin2024vqascore} uses a Yes-answer
probability for task alignment. These are adaptations of existing evaluators
to the available prediction interface, not evaluations on a generated temporal
rollout. CLIP based components use
CLIP~\citep{radford2021clip},
\ifdefined\WAMArxivRelease\else
the WMReward representation score uses V-JEPA 2~\citep{assran2025vjepa2},
\fi
and
structural similarity scores use SSIM~\citep{wang2004ssim}.

These input adaptations differ from the image-pair scoring used in the matched
same state study below. Per decision records for the repeated
frame evaluation are unavailable, so its
reported success aggregates cannot establish the realised tie rates or all
runtime score directions. The decoded slot evaluation retains per run summaries and input
manifests. Its summaries do not count scorer failures, and some VLM parsers
return zero on a failed response. We therefore distinguish observed score ties
from successful, informative judgements. We report success and its change from
the $K=1$ reference, with ranges spanning the 2.5 and 97.5 percentiles of 10,000
resamples of the 24 tasks.
Resampling tasks preserves their trials and exposes sensitivity to task composition.

\subsection{What the Metrics Recover}

The separately run $K=1$ reference succeeds on $65.9\%$ of episodes.
As shown in \cref{tab:closed_loop_by_difficulty}, its success ranges from
$31.0\%$ on hard tasks to $94.3\%$ on easy tasks. Some selectors yield higher
observed success. With $K=4$, learned value selection reaches $66.8\%$
overall and background consistency reaches $66.7\%$. On medium tasks,
background consistency reaches $69.3\%$, compared with $66.9\%$ for the
reference. Learned value has a higher point estimate in all three groups,
whereas background consistency loses success on easy tasks. The task resampled
ranges for their overall changes are $-1.46$ to $+3.17$ and $-1.58$ to $+3.17$
percentage points, respectively. Both ranges include negative and positive
changes, so they neither establish a gain nor rule out a moderate improvement.
\ifdefined\WAMArxivRelease\else
WMReward and Robometer, omitted from the
main table, reach $61.8\%$ and $60.0\%$, respectively.
\fi
These results measure selection through the
adapted slot interfaces, not evaluation on a predicted temporal rollout.
Across all 24 tasks, VLM controllability and VLM progress value assign
identical scores to every candidate at $60.4\%$ and $79.1\%$ of decisions,
respectively. Ties select the
first candidate. These rules therefore often leave the candidate choice
unchanged, limiting what their aggregate success reveals about visual ranking.
\ifdefined\WAMArxivRelease\else
Neither WMReward nor Robometer assigns identical scores to all candidates at
any recorded decision.
\fi

Because the $K=1$ reference is not drawn from each saved $K=4$ set, this
online comparison does not measure how much of the local selection headroom a selector
recovers. It instead measures end to end performance under each rule.
Aggregate success alone does not identify where an evaluator gains or loses
performance. We therefore use same state candidate sets to measure where
successful alternatives exist and whether the scores select them.

\section{Understanding the Selection Gap}
\label{sec:selection_gap}

\begin{figure*}[t]
\centering
\includegraphics[width=0.90\linewidth]{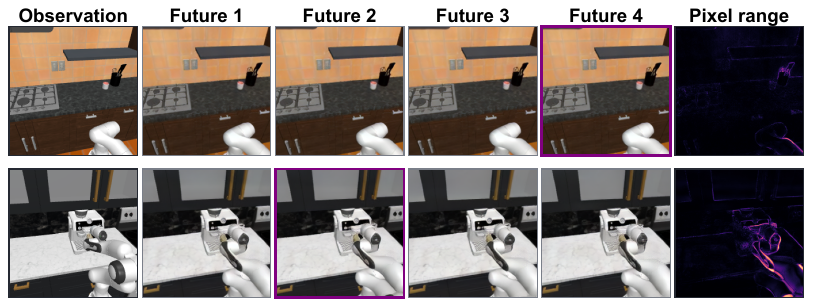}
\caption{\textbf{Visual evidence available to the selector.}
  Each row shows the current observation and four sampled visual futures. The
  upper example is closest to the median pairwise pixel error among 194
  decisions, while the lower example has the largest error. Their errors are
  $0.44\%$ and $1.96\%$. Purple borders indicate the candidate preferred by the
  learned value and do not denote task outcome.}
\label{fig:candidate_examples}
\end{figure*}

A selector can fail because the candidate actions reach the same outcome,
because their predicted futures do not expose the relevant difference, or
because the score ranks that evidence incorrectly. We first measure outcome
differences and ranking on counterfactual branches, then examine what the
visual feature probes can establish. In \cref{fig:candidate_examples}, two
recorded candidate sets from the quantitative spread panel illustrate the
input available to a selector. The branch experiments separately test whether
sampled actions differ in terminal success.

\subsection{Task Consequence Diversity}

The 4-task same state experiment records all candidate outcomes, separating
how often a choice matters from how difficult that choice is.
Let $M$ be the set of decision states with both successful and failed candidate
continuations, and let $q_t$ be the fraction of successful candidates at state
$t$. For $N$ recorded states, the gain available over uniform selection is
\begin{equation}
H = \frac{1}{N}\sum_{t\in M}(1-q_t).
\label{eq:local_headroom}
\end{equation}
States where all candidates succeed or all fail contribute nothing.
Where both outcomes occur, rare successes are harder to find by uniform
selection. Both state frequency and success balance therefore matter.

\begin{figure}[t]
\centering
\includegraphics[width=\columnwidth]{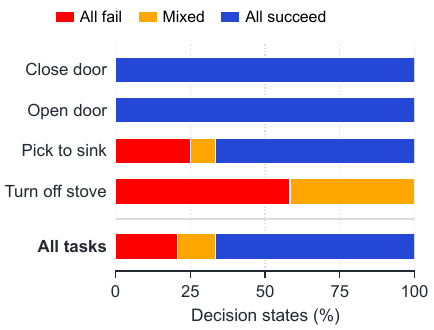}
\caption{\textbf{Selection can change success only at mixed decisions.}
  This 4-task experiment uses the first 4 replanning decisions
  from 3 trajectories per task.
  Each decision has 16 candidate continuations. Mixed decisions contain both
  successful and failed outcomes. Only 6 of the 48 decisions are mixed.}
\label{fig:branch_outcome_composition}
\end{figure}

In the initial $K=16$ experiment, the 6 states with both outcomes have only
$17.7\%$ successful candidates on average. They account for the full difference
between uniform and oracle selection. The remaining 42 states provide
no terminal success advantage to any ranking rule under the recorded
continuation. As shown in \cref{fig:branch_outcome_composition}, these outcomes
are unevenly distributed across the 4 tasks. The 2 door tasks contain only successful continuations,
whereas the pick and place and stove tasks contain both uniformly failed and
mixed decisions. Selection opportunity is therefore uneven even within a task.

In a separate subset comparison, greedy farthest point sampling selects $K=4$
or $K=8$ candidates from all 32 archived samples at each of the same 48 states.
It favours candidates whose first 16 actions are farther apart in action space.
As shown in \cref{tab:action_spread_subsets},
this action spread increases relative to the first $K$ candidates, but oracle
success does not. The fraction of states with both outcomes is also unchanged at
$6.3\%$ for $K=4$ and $10.4\%$ for $K=8$. Uniform success decreases, so the
gap between oracle and uniform success grows without any increase in oracle success.
Thus, the recorded subset rule changes numerical spread and success balance,
but does not find successful alternatives at more states than the ordered
subsets. Against 1,000 random subset draws, its oracle success is slightly
higher, by $0.1\%$ at $K=4$ and $1.0\%$ at $K=8$. This conditional comparison
reuses an existing candidate set rather than testing a broader generator.

\begin{table}[t]
\centering
\caption{\textbf{Action spread and oracle coverage.}
  Subsets from 32 archived candidates at each of 48 decision states. Random rows
  average 1,000 draws from the same candidate sets.
  Uniform and oracle columns give success percentages. Gap is their difference
  in percentage points. Spread measures pairwise differences between the
  executed action sequences.}
\label{tab:action_spread_subsets}
\footnotesize
\setlength{\tabcolsep}{2.4pt}
\renewcommand{\arraystretch}{1.05}
\begin{tabular*}{\columnwidth}{@{\extracolsep{\fill}}clrrrr@{}}
\toprule
$K$ & \textbf{Subset} & \textbf{Spread} & \textbf{Uniform} & \textbf{Oracle} & \textbf{Gap} \\
\midrule
4 & Ordered & 0.00649 & 69.27 & 72.92 & 3.65 \\
  & Random & 0.00631 & 68.64 & 72.85 & 4.22 \\
  & Wider spread & 0.00880 & 68.23 & 72.92 & 4.69 \\
\midrule
8 & Ordered & 0.00633 & 69.01 & 77.08 & 8.07 \\
  & Random & 0.00630 & 68.63 & 76.11 & 7.48 \\
  & Wider spread & 0.00764 & 68.49 & 77.08 & 8.59 \\
\bottomrule
\end{tabular*}
\end{table}

\subsection{Ranking Where Outcomes Differ}

We next ask whether scores identify successful candidates where the choice
matters. We score the same 16 candidates per state before loading their outcomes.
The three prespecified signals are learned value, greyscale SSIM
between the current image and final predicted frame, and VLM progress.
The latter uses a fixed Qwen3-VL image-pair prompt and the normalised
probability of Yes rather than a parsed numerical rating. All 768 scores
are finite and matched to their candidates. SSIM measures similarity to the
current image, not accuracy against the realised future, whose frames are unavailable.

At $K=16$, learned value and VLM progress each select successful candidates
at 32 of 48 states, giving $66.7\%$. SSIM selects 33, giving
$68.8\%$. All three remain below the $68.9\%$ uniform expectation. Only the 6
states with both outcomes distinguish ranking rules. There, learned
value and VLM progress select no successful candidate, while SSIM selects one.
Uniform selection would yield $1.06$ successful choices in expectation across
these 6 states, so these counts do not establish below-random performance.
Within state outcome AUC measures whether a successful candidate is
scored above a failed candidate, counting a tie as one half. Averaging across these
states gives $0.54$ for learned value and $0.40$ for both SSIM and VLM progress,
compared with $0.5$ for random ranking.

\ifdefined\WAMExtendedVersion
To examine more than the top choice, we retain the $r$ highest ranked candidates
from each fixed set of 16 and measure the fraction of mixed decisions whose
retained set contains a successful candidate. Retaining four
candidates gives $66.7\%$ under learned value, compared with $56.9\%$ in
expectation under random ranking. At eight retained candidates, these values
are $100\%$ and $85.4\%$. Retaining successful alternatives is therefore not
by itself evidence of useful scoring. The comparison with random ranking
matters, especially as the retained set grows.

At one mixed state, VLM progress assigns a successful candidate the joint highest
score, but the fixed tie rule chooses a failed candidate. Since the 6 mixed
states arise from only 3 behaviour trajectories, these results describe
the observed cases rather than establish a reliable ranking advantage. They
also concern the contents of a retained set, not a controller that can identify
and execute its successful member.

\begin{figure}[t]
\centering
\includegraphics[width=\columnwidth]{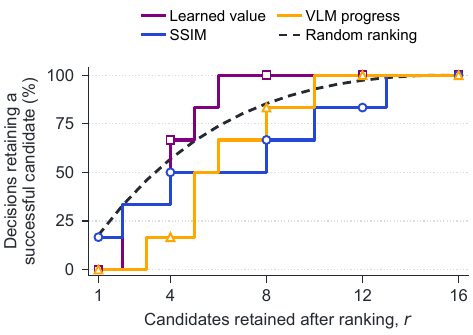}
\caption{\textbf{Ranking compared with random candidate retention.}
  For the 6 mixed decisions in \cref{fig:branch_outcome_composition}, curves
  show whether the top $r$ candidates contain a success. The candidate set
  stays fixed at 16 candidates. The dashed curve is the exact expectation under
  random ranking. Score ties use the recorded candidate order. Retention is not
  executed task success.}
\label{fig:ranking_retention}
\end{figure}
\else
Keeping the four highest ranked candidates retains a success at $66.7\%$ of
mixed states under learned value, compared with $56.9\%$ under random ranking.
Keeping eight gives $100\%$ and $85.4\%$, respectively. This set coverage is
not executed success, and grows even under random ranking. At one mixed state,
VLM progress ties a successful branch for the highest score but chooses a
failure under the fixed index rule. The six mixed cases come from only three
trajectories, limiting conclusions about a reliable ranking advantage.
\fi

\subsection{Full Chunk Selection Across Trajectory Phases}
\label{sec:full_chunk}

We test whether the selection gap persists beyond early decisions and when the
complete predicted action sequence is executed. A separate experiment
samples one eligible decision uniformly from each of 16 episodes per
task, using the same 4 tasks. Its 64 states span early, middle and late
trajectory phases, with 16 fresh candidates each. Both behaviour and candidate
continuation execute all 32 predicted actions before replanning, matching the
initial execution to the future image horizon. Candidate seeds and scores are
fixed before outcomes are joined. All 1,024 continuations reach success or the task
limit. We also evaluate the first four candidates as a nested comparison.

In \cref{tab:full_chunk_selection}, the outcome oracle reaches $59.4\%$ at
$K=16$, compared with $50.2\%$ under uniform selection. Learned value reaches
$51.6\%$, whereas SSIM and VLM progress each reach $50.0\%$. Thus, learned
value provides a small observed gain but leaves most of the opportunity
unrecovered. Random tie resolution reduces its success to $50.3\%$.
Paired episode resampling within each task includes zero for
every selector's change from uniform, while the remaining oracle gaps stay
positive. There are 13 states with both outcomes, distributed across all 4
tasks. The gap therefore persists beyond the early decisions in the initial
panel, and when complete chunks are executed. Since both the sampled states
and execution horizon change, this is not an isolated test of either factor.
\ifdefined\WAMExtendedVersion
Task level results, uncertainty and tie sensitivity appear in
\cref{app:full_chunk}.
\fi

\begin{table}[t]
\vspace*{5pt}
\centering
\caption{\textbf{Selection with complete action chunks.}
  The same 64 decision states are evaluated at both candidate counts.
  Success is measured after a common continuation. $\Delta$ is the
  change from uniform selection at $K=16$, in percentage points.}
\label{tab:full_chunk_selection}
\footnotesize
\renewcommand{\arraystretch}{1.05}
\begin{tabular*}{\columnwidth}{@{\extracolsep{\fill}}lrrr@{}}
\toprule
\textbf{Selection rule} & $K=4$ & $K=16$ & $\Delta$ \\
\midrule
\rowcolor{tablegray}[0pt][0pt]Uniform & 50.4 & 50.2 & +0.00 \\
\rowcolor{tablegray}[0pt][0pt]Learned value & 50.0 & 51.6 & \textcolor{metricgreen}{+1.37} \\
SSIM & 50.0 & 50.0 & -0.20 \\
VLM progress & 50.0 & 50.0 & -0.20 \\
\midrule
Outcome oracle & 54.7 & 59.4 & +9.18 \\
\bottomrule
\end{tabular*}
\end{table}

\subsection{What the Visual Diagnostics Measure}
\label{sec:native_video}

\ifdefined\WAMExtendedVersion
Across 194 decisions, mean pairwise pixel error between predicted futures is
$0.60\%$. Across 130 decisions, the median spread in learned value is
$0.002$. The five recorded scores in a
separate set of 2,400 recorded demonstration states have within decision variation below
$0.06$ of their variation across the full panel. Three average the current
image together with the predictions, so their ratios describe the saved input
rather than variation in predicted images alone. These measurements do not
establish whether the small differences are relevant to task outcome.

\ifdefined\WAMSubmissionRevision
\else
\begin{figure}[t]
\centering
\includegraphics[width=\columnwidth]{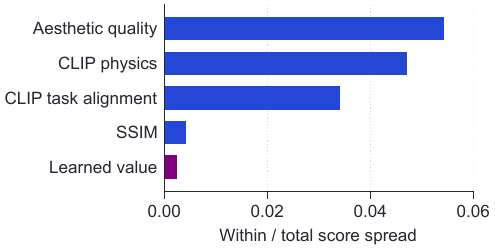}
\caption{\textbf{Within decision spread of recorded scores.}
  Bars show mean within decision standard deviation relative to the full
  set of 2,400 demonstration states. Aesthetic quality, CLIP physics and CLIP task alignment
  average current and predicted images. Spread does not measure task outcome
  discrimination.}
\label{fig:candidate_spread}
\end{figure}
\fi

Small score differences need not be uninformative. What matters is whether they
order the candidate actions correctly. The terminal outcomes above test that
ordering directly. A separate historical feature probe instead predicts which
action chunk is closest to a recorded demonstration. Matching a demonstrator
is a different target from recognising a successful alternative, so that probe
does not resolve the source of the selection gap.

The 194-decision visual panel supports a second, distinct diagnostic. Every
candidate prediction is compared with the same observation, recorded after the
first action of candidate zero. This measures agreement with one behaviour
transition, rather than the realised consequence of each candidate. Correcting
the scores' current-image reference and frame reductions yields $60.5\%$ gap
recovery for SSIM. This corresponds to only a $0.25\%$ reduction in pixel error
relative to uniform selection. Under its recorded frame reduction, CLIP task
alignment instead increases this error, with gap recovery of $-30.8\%$.
Removing the shared current image from the aesthetic quality, CLIP physics
and CLIP task alignment averages preserves every candidate ranking in this
panel. Their pixel error recoveries therefore remain unchanged.
Similarity to the current image can
therefore rank this short-horizon proxy without establishing task progress.
\else
Across 194 decisions, mean pairwise pixel error between predicted futures is
$0.60\%$. The median learned value spread across 130 decisions is $0.002$.
Small differences need not be uninformative, but their magnitude alone does
not establish whether they predict task outcome. A historical feature probe
instead targets the action closest to a demonstration, which is not the same
as recognising a successful alternative.

The visual panel compares every candidate with one shared observation after
the first action of candidate zero. Correcting the current image reference
and frame reductions lets SSIM recover $60.5\%$ of the available pixel error
gap, but reduces error by only $0.25\%$ relative to uniform selection. Recorded
CLIP task alignment instead increases error. Removing the current image from
three scores that average current and predicted images leaves every ranking
unchanged. These results concern a shared behaviour transition, not the
realised consequence of each candidate or task progress.
\fi

These diagnostics distinguish image agreement from action utility, but do not
identify whether a missed success is obscured by the prediction or overlooked
by the score. The full chunk evaluation matches the initial execution to the
prediction horizon, but does not retain the corresponding realised images.
Testing what evidence is missing therefore still requires predicted and
realised observations for the same candidate actions at the same horizon.
\ifdefined\WAMExtendedVersion
Additional diagnostics appear in \cref{app:complete_selector_results}.
\fi

\ifdefined\WAMSubmissionRevision
\section{Additional Model Checks}
\else
\section{Additional Model and Sampler Checks}
\fi
\label{sec:additional_evidence}

We also examine how selection behaves with additional models and benchmarks.
\ifdefined\WAMSubmissionRevision
The additional checks change benchmark and action interface as well as model,
so they do not replicate the same state outcome study. Learned value selection
with Cosmos Policy on four LIBERO~\citep{liu2023libero} suites changes success
from $98.05\%$ to $98.00\%$. RynnVLA~\citep{cen2025rynnvla} reaches $85.6\%$
with one sample and $85.0\%$ under random selection from four candidates.
LingBot-VA~\citep{li2026lingbotva} reaches $97.0\%$ with one sample and
$97.5\%$ under both random and VLM selection from four candidates. The equal
random and VLM estimates do not isolate a scoring benefit, and the high success
rates leave little room for improvement. Further matched outcome comparisons
are needed to test whether the primary diagnosis transfers across WAMs.
\else
\paragraph{Within model controls}
Changing the Cosmos sampler across four settings leaves within decision action
variance below $0.005$ of marginal variance. Since these interventions change
spread only weakly, they do not test a substantially more diverse generator.
The cross interface check in \cref{fig:transfer} compares relative action
spread across three systems. It measures the sampled actions directly and does
not depend on a visual scorer. Since the model, benchmark and action interface
change together, it is a descriptive breadth check rather than a matched
replication of the outcome experiment.

\begin{figure}[t]
\centering
\includegraphics[width=\columnwidth]{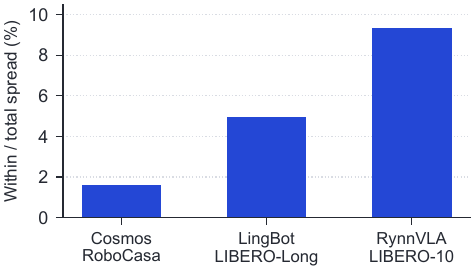}
\caption{\textbf{Action spread across WAM interfaces.}
  Bars show within decision action spread as a fraction of marginal spread.
  Each system uses its compatible benchmark. This statistic measures numerical
  action differences, not differences in task outcome.}
\label{fig:transfer}
\end{figure}

\paragraph{Across model breadth}
As shown in \cref{fig:transfer}, the same dimensionless action statistic gives
within decision to marginal
spread ratios of $0.016$, $0.049$, and $0.093$ for Cosmos Policy,
LingBot-VA~\citep{li2026lingbotva}, and RynnVLA~\citep{cen2025rynnvla}.
Compatible closed loop checks produce only
small changes whose direction depends on the model and setting. Learned value
selection with Cosmos Policy on four
LIBERO~\citep{liu2023libero} suites changes
success from $98.05\%$ to $98.00\%$. RynnVLA reaches $85.6\%$ with one sample
and $85.0\%$ under random selection from four candidates. LingBot-VA reaches
$97.0\%$ with one sample and $97.5\%$ under both random and VLM based selection
from four candidates. The identical random and VLM estimates do not isolate a
benefit from scoring, and near ceiling success limits measurable headroom.
\ifdefined\WAMExtendedVersion
Additional transfer analyses appear in \cref{app:transfer}.
\fi
\fi

\section{Discussion}
\label{sec:discussion}

\textbf{Generating useful actions.} A WAM should complete tasks reliably,
not pursue diversity for its own sake. Ranking cannot help when all candidates
succeed or all fail. Where both outcomes occur, rare successful candidates
offer more improvement over uniform selection. Numerical action differences
alone do not establish whether these useful choices exist.

\textbf{Recognising useful alternatives.} Some scores yield higher observed
online success, but matched ranking often misses successful alternatives.
Retaining a successful action among several highly ranked candidates does not
mean a controller can identify it. In the full chunk panel, learned value
gives a small observed gain but remains well below the oracle. Greater action
spread leaves oracle success unchanged relative to ordered subsets, while
slightly exceeding the average random subset. Together, these
analyses separate the availability of useful choices from their recognition.

\textbf{Implications for WAM evaluation.} Executed outcomes complement visual quality by measuring
the opportunity that generation provides and the fraction a selector recovers.
Determining whether a missed alternative is obscured by the prediction or overlooked by
the score requires comparing predicted and realised evidence for the same
actions at matched horizons. Proposal expansion~\citep{kwok2026cover}, future
consensus~\citep{ruan2026futurecompatible}, and geometric
verification~\citep{zhao2026geobon} provide complementary approaches to stronger
selection. Reporting available opportunity alongside their recovered gain
would help explain when different approaches succeed.

\textbf{Scope and next steps.} The controlled evidence concerns Cosmos Policy in RoboCasa under a fixed
continuation. The full chunk panel extends the analysis across trajectory
phases with one decision state per episode, but changes the execution horizon as well.
It supports the remaining selection gap, not a causal explanation for changes
in mixed decision frequency. A matched second WAM remains a useful next test
because the broader model checks change both benchmark and interface.

\section{Conclusion}

In this paper, we studied test time planning with world action models by
separating available selection opportunity from the gain recovered by scoring
their generated futures. In the initial candidate sets,
this opportunity is concentrated in relatively few decisions, and tested
scores often miss their successful alternatives. The full chunk evaluation
across trajectory phases also leaves substantial opportunity unrecovered,
despite a small gain from learned value. These findings distinguish producing
useful action choices from recognising them, motivating the evaluation of WAM
predictions through their usefulness for decisions alongside their visual
quality.

\ifdefined\WAMArxivRelease
  \section*{Acknowledgements}
This research was supported by the NVIDIA Academic Grant Program using simulation and compute resources and was supported by the EPSRC Programme Grant ``From Sensing to Collaboration'' (EP/V000748/1).

\else
  \ifdefined\WAMSubmissionRevision
    \input{sections/acknowledgements_ai}
  \fi
\fi

\balance
\bibliographystyle{IEEEtran}
\bibliography{references}

\ifdefined\ICRASubmissionOnly
\else
  \clearpage
  \nobalance
  \appendices
  \section{Extended Experimental Results}
\label[appendix]{app:complete_record}

We report the opportunity protocols and matched ranking analyses, followed by
the reported selector panels, visual diagnostics, and additional sampler and
model checks. Historical exploratory analyses are identified separately.

\begin{table*}[t]
\vspace*{5pt}
\centering
\caption{\textbf{Evidence panels and their targets.} Panel sizes refer to the
  recorded evaluation units, not interchangeable independent observations.}
\label{tab:panel_map}
\footnotesize
\renewcommand{\arraystretch}{1.1}
\begin{tabular*}{\textwidth}{@{\extracolsep{\fill}}lll@{}}
\toprule
\textbf{Panel} & \textbf{Recorded units} & \textbf{Target and interpretation} \\
\midrule
Trajectory coverage & 24 tasks, 720 scenes, 16 seeds & Terminal success of complete policy runs \\
Online selection & 24 tasks, 50 trials per rule and task & Executed control under each rule, not matched local recovery \\
Frozen branches & 12 trajectories, 48 anchors, 16 candidates & Candidate terminal outcomes under a common continuation \\
Full chunk branches & 64 episodes, 64 anchors, 16 candidates & Phase sampled decisions with 32 action execution \\
Action-spread subsets & Same 48 anchors, archived pool of 32 & Outcome coverage within selected subsets \\
Visual proxy & 194 decisions, 12 task and trial groups & Agreement with one shared behaviour-transition image \\
Value spread & 130 decisions & Within-decision variation, not task-outcome discrimination \\
Historical feature probe & 240 trajectories, 2,400 demonstration anchors & Closest demonstration action, not executed success \\
\bottomrule
\end{tabular*}
\end{table*}
The learned value online condition pools two independently run panels and
therefore has 100 trials per task. In \cref{tab:panel_map}, we separate data sources that do
not share candidate-specific terminal labels.

\subsection{Prediction and Execution Interface}
\label[appendix]{app:execution_interface}

\begingroup\raggedright
The recorded primary checkpoint is
\path{nvidia/Cosmos-Policy-RoboCasa-Predict2-2B}, using the
\path{cosmos_predict2_2b_480p_robocasa_50_demos_per_task__inference}
configuration.\par
\endgroup
Each sampled sequence contains 32 predicted actions. The
controller executes its first 16 actions before requesting another prediction
in the primary panels. The separate full chunk panel executes all 32 actions.
We use $\hat a^{(i)}$ throughout the selection notation for the executable
part of the sequence. Each branch evaluation executes this part and then follows
its fixed continuation, with a common seed schedule across candidates.

The future image target in Cosmos Policy corresponds to the state after the
predicted action sequence. Four copies of the image at that timestep are
encoded into one future camera slot~\citep{cosmospolicy2026}. Thus, the four
decoded slot images are not predictions at four successive times. Moreover,
the branch outcome after executing a prefix and continuing the policy is not
a direct fidelity test at the full predicted horizon. The branch oracle
evaluates the deployed action choice, whereas the image diagnostics evaluate
the visual evidence available to the selector.

Our fixed-checkpoint study does not reproduce the separate planning system in
Cosmos Policy that refines a world model and value function with policy rollout
data~\citep{cosmospolicy2026}. This distinction separates an analysis of the
released paired predictions from evaluation of that refined planning system.

\subsection{Complete Opportunity Analysis}
\label[appendix]{app:opportunity}

\paragraph{Complete trajectory coverage}
The matched trajectory bank contains 24 tasks, 30 scenes per task, and 16
policy noise seeds per scene. All 11,520 trajectories are complete. The scene
identity is a deterministic function of the base seed and episode index. The
policy seed offset is the only quantity varied within a scene. Exact ordered
prefixes yield pooled success counts of 463, 492, 514, 538, and 557 out of 720
instances for $K=1,2,4,8,16$. The corresponding whole suite pass@$K$ values
are $64.3\%$, $68.3\%$, $71.4\%$, $74.7\%$, and $77.4\%$. Because every
task contributes 30 scenes, the average over tasks and pooled average coincide.
Task resampled intervals are $53.8\%$ to $74.2\%$ at $K=1$ and
$69.2\%$ to $84.9\%$ at $K=16$.
Because exact prefixes depend on seed order, we also use the standard
combinatorial estimator that averages over all subsets of size $K$ from each
16-seed bank. It gives $64.4\%$, $69.0\%$, $72.6\%$, $75.3\%$, and $77.4\%$
for $K=1,2,4,8,16$, closely matching the ordered prefix curve.
In \cref{tab:passk_task_level}, we report the results for every task.

\begin{table*}[t]
\vspace*{5pt}
\centering
\caption{\textbf{Task level complete trajectory pass@$K$ (\%).}
  Each task contains 30 environment matched scenes and 16 ordered policy
  seeds. Values use exact nested prefixes of the same trajectory bank.}
\label{tab:passk_task_level}
\footnotesize
\setlength{\tabcolsep}{2.1pt}
\renewcommand{\arraystretch}{0.92}
\begin{tabular}{@{}lrrrrr@{\qquad}lrrrrr@{}}
\toprule
\textbf{Task} & \textbf{1} & \textbf{2} & \textbf{4} & \textbf{8} & \textbf{16} &
\textbf{Task} & \textbf{1} & \textbf{2} & \textbf{4} & \textbf{8} & \textbf{16} \\
\midrule
CloseDoubleDoor       & 86.7 & 90.0 & 90.0 & 93.3 & 96.7 & PnPCounterToSink      & 56.7 & 70.0 & 73.3 & 73.3 & 73.3 \\
CloseDrawer           &100.0 &100.0 &100.0 &100.0 &100.0 & PnPCounterToStove     & 53.3 & 56.7 & 63.3 & 66.7 & 73.3 \\
CloseSingleDoor       &100.0 &100.0 &100.0 &100.0 &100.0 & PnPMicrowaveToCounter & 30.0 & 33.3 & 40.0 & 46.7 & 50.0 \\
CoffeePressButton     & 86.7 & 90.0 & 90.0 &100.0 &100.0 & PnPSinkToCounter      & 63.3 & 66.7 & 73.3 & 76.7 & 80.0 \\
CoffeeServeMug        & 53.3 & 56.7 & 56.7 & 63.3 & 66.7 & PnPStoveToCounter     & 50.0 & 63.3 & 66.7 & 70.0 & 70.0 \\
CoffeeSetupMug        & 20.0 & 23.3 & 33.3 & 50.0 & 63.3 & TurnOffMicrowave      & 96.7 &100.0 &100.0 &100.0 &100.0 \\
OpenDoubleDoor        & 86.7 & 90.0 & 93.3 & 96.7 & 96.7 & TurnOffSinkFaucet     & 70.0 & 83.3 &100.0 &100.0 &100.0 \\
OpenDrawer            & 86.7 & 90.0 & 90.0 & 93.3 & 93.3 & TurnOffStove          & 10.0 & 10.0 & 10.0 & 20.0 & 26.7 \\
OpenSingleDoor        & 86.7 & 86.7 & 86.7 & 86.7 & 86.7 & TurnOnMicrowave       & 56.7 & 60.0 & 63.3 & 66.7 & 70.0 \\
PnPCabToCounter       & 43.3 & 50.0 & 53.3 & 53.3 & 53.3 & TurnOnSinkFaucet      & 70.0 & 73.3 & 73.3 & 76.7 & 80.0 \\
PnPCounterToCab       & 63.3 & 63.3 & 63.3 & 63.3 & 63.3 & TurnOnStove           & 56.7 & 66.7 & 66.7 & 66.7 & 73.3 \\
PnPCounterToMicrowave & 26.7 & 26.7 & 36.7 & 40.0 & 50.0 & TurnSinkSpout         & 90.0 & 90.0 & 90.0 & 90.0 & 90.0 \\
\midrule
\textbf{Task mean} & \textbf{64.3} & \textbf{68.3} & \textbf{71.4} & \textbf{74.7} & \textbf{77.4} &
& & & & & \\
\bottomrule
\end{tabular}
\end{table*}

\paragraph{Same state coverage}
The fixed candidate panel contains 12 behaviour trajectories and 48 recorded
decision states, termed anchors below. We report the first 16 sampled
candidates and their continuations per anchor. Candidate zero determines the behaviour
trajectory from which anchors are captured. Counterfactual branches cannot
alter later anchors. The snapshot includes simulator and model arrays,
controller and observable buffers, and random number state. Restored hashes
match at all 1,536 archived branches, and all 48 anchors pass the prefix replay
isolation check. The continuation policy uses the same seed schedule across
candidates.

At $K=16$, the mean branch, best branch, and mixed anchor rates are $68.9\%$,
$79.2\%$, and $12.5\%$. The two door tasks remain saturated at every reported
prefix. The mixed anchors occur in the other two manipulation tasks. A matched
$K=16$ comparison scores these exact branches before loading their terminal
outcomes. Learned value selects 32 successful branches out of 48,
SSIM selects 33, and VLM progress selects 32. Their success
rates are $66.7\%$, $68.8\%$, and $66.7\%$, compared with a $68.9\%$ uniform
expectation and $79.2\%$ best candidate success. Across the six mixed anchors,
branch outcome AUC is $0.54$, $0.40$, and $0.40$, respectively. SSIM measures
anchoring to the current frame rather than realised prediction fidelity. A
physics score is not reported because the frozen bank lacks realised branch
frames, contacts, poses, and the clip length required by the available metric.
The six mixed anchors arise from three behaviour trajectories. Their individual
outcomes appear in \cref{tab:mixed_anchor_cases}. The full decision map in
\cref{fig:branch_outcome_cases} retains the trajectory identities aggregated in
\cref{fig:branch_outcome_composition}.

\begin{table}[t]
\centering
\caption{\textbf{Selections on the six mixed $K=16$ anchors.}
  The candidate column gives successful branches out of 16. A value of one in a selector
  column means that its chosen branch succeeds.}
\label{tab:mixed_anchor_cases}
\scriptsize
\setlength{\tabcolsep}{2.3pt}
\renewcommand{\arraystretch}{0.94}
\begin{tabular*}{\columnwidth}{@{\extracolsep{\fill}}llrrrr@{}}
\toprule
\textbf{Task} & \textbf{Anchor} & \textbf{Candidates} & \shortstack{\textbf{Learned}\\\textbf{value}} &
\textbf{SSIM} & \shortstack{\textbf{VLM}\\\textbf{progress}} \\
\midrule
Pick counter to sink & E0--D1 & 2/16 & 0 & 0 & 0 \\
Turn off stove & E0--D0 & 4/16 & 0 & 0 & 0 \\
Turn off stove & E0--D1 & 2/16 & 0 & 0 & 0 \\
Turn off stove & E0--D2 & 4/16 & 0 & 1 & 0 \\
Turn off stove & E0--D3 & 3/16 & 0 & 0 & 0 \\
Turn off stove & E2--D3 & 2/16 & 0 & 0 & 0 \\
\bottomrule
\end{tabular*}
\end{table}

\paragraph{Principal matched scorer interface}
The frozen VLM uses \path{Qwen/Qwen3-VL-4B-Instruct} at revision
\path{ebb281ec70b05090aa6165b016eac8ec08e71b17}. Its left image is the current
primary-camera observation and its right image is frame index 3 from the
candidate's decoded primary-camera future slot. The fixed prompt asks whether
the task is more complete in the right image than in the left. The score is
the normalised Yes probability from the first Yes and No token logits, with
deterministic inference. This is not a generated numerical rating. The scorer
specification binds the model files, input images and candidate identities
before outcomes are joined. Missing and nonfinite scores fail validation.
Higher scores are preferred, with ties resolved by ascending candidate index.
These recorded controls concern the matched K16 analysis, not the separate
legacy online panel with unavailable per-decision logs.

\begin{figure}[t]
\centering
\includegraphics[width=\columnwidth]{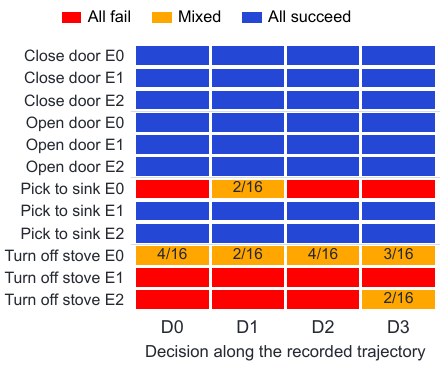}
\caption{\textbf{Individual decisions in the four-task branch panel.}
  Each cell summarises 16 candidate outcomes. Rows identify behaviour
  trajectories E0 to E2 and columns give their first four recorded replanning
  decisions. Numbers in mixed cells count successful branches. This is not a
  map of decision sensitivity over complete trajectories.}
\label{fig:branch_outcome_cases}
\end{figure}

\paragraph{Ranking successful alternatives}
In \cref{fig:ranking_retention}, we sort candidates by descending score,
breaking ties by ascending candidate index as in the deposited selection
analysis. At a mixed anchor, the first successful rank is the smallest retained
set that contains a successful branch. Learned value retains a successful
branch at zero, two, four and six anchors when keeping its top 1, 2, 4 and 8
candidates. The corresponding counts are one, two, three and four for SSIM,
and zero, zero, one and five for VLM progress. All counts use the same six
mixed banks. They describe a retained set, not the performance of a controller
that can choose correctly within it. For a bank of $K$ candidates containing
$s$ successes, a uniform random ranking retains a success in its first $r$
positions with probability
\begin{equation}
1 - \frac{\binom{K-s}{r}}{\binom{K}{r}}.
\end{equation}
We take $\binom{K-s}{r}=0$ when $r>K-s$ and average this probability over
the six mixed banks at fixed $K=16$. This exact finite-bank expectation gives
$17.7\%$, $32.9\%$, $56.9\%$ and $85.4\%$ at $r=1,2,4,8$. It requires
neither additional rollouts nor randomly sampled permutations.

Only VLM progress has tied scores in these six banks. For its highest scoring
successful group, let $a$ candidates have strictly higher scores, and let the
tied group contain $n$ candidates of which $s$ succeed. The first successful
rank can range from $a+1$ to $a+n-s+1$ across tie orderings.
In \cref{fig:successful_candidate_ranks}, markers show the recorded index
ordering and horizontal lines display this range. The main retention curve uses
that same recorded ordering.

\begin{figure}[t]
\centering
\includegraphics[width=\columnwidth]{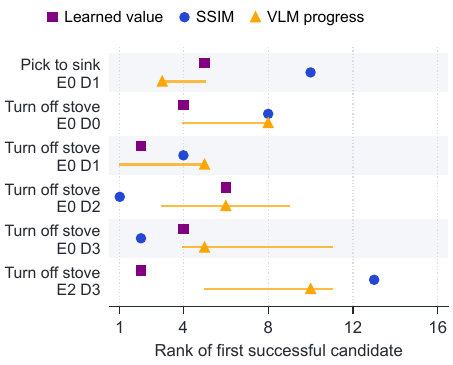}
\caption{\textbf{Rank of the first successful candidate at each mixed decision.}
  Rows correspond to the six mixed cells in \cref{fig:branch_outcome_cases}.
  Rank 1 is the selector's choice. Markers use the recorded candidate order to
  break score ties. Horizontal lines span the possible first successful ranks
  under other orders of tied scores, not statistical uncertainty.}
\label{fig:successful_candidate_ranks}
\end{figure}

\paragraph{Trajectory sensitivity}
We repeat the local headroom calculation after excluding each of the 12
behaviour trajectories in turn, retaining 44 anchors each time. The available
gain ranges from $4.0\%$ to $11.2\%$, compared with $10.3\%$ on all 48 anchors.
The largest reduction occurs when excluding trajectory E0 of
\texttt{TurnOffStove}, which contains four of the six mixed decisions. This
descriptive sensitivity analysis preserves the dependence between decisions
on one trajectory and shows how strongly the magnitude depends on the few
trajectories that contain consequential choices.

\paragraph{Action spread subsets}
The historical subset comparison selects from all 32 archived candidates at
each of the same 48 anchors, not exclusively from the first 16 candidates used
by the matched scoring study. The action spread rule chooses $K=4$ or $K=8$
candidates. The deposited mean action-spread statistic rises from $0.006493$
to $0.008796$ at $K=4$, and from $0.006325$ to $0.007636$ at $K=8$.
Its subsets match the ordered prefixes in mixed rate and best
branch success. These are $6.3\%$ and $72.9\%$ at $K=4$, and $10.4\%$ and
$77.1\%$ at $K=8$. Uniform success changes from $69.27\%$ to $68.23\%$ at
$K=4$ and from $69.01\%$ to $68.49\%$ at $K=8$. Consequently, headroom
increases from 3.65 to 4.69 percentage points and from 8.07 to 8.59 points,
respectively, without higher oracle success. As shown in
\cref{tab:action_spread_subsets}, more spread changes the success balance but
not the fraction of mixed anchors relative to the ordered subsets.
The recovered construction uses greedy farthest point sampling on the raw
flattened $16\times7$ action prefixes in double precision, starting with the
candidate farthest from the centroid. It applies no coordinate scaling. The
reported spread is the mean pairwise Euclidean distance divided by $\sqrt{112}$.
Reapplying the recovered implementation to the archive reproduces all 192
recorded subsets and their spread statistics.

We additionally compare 1,000 uniform random subsets at each candidate count,
using master seed 195 with a separate stream for each candidate count and the
same saved banks. Their average oracle success is
$72.85\%$ at $K=4$ and $76.11\%$ at $K=8$, compared with $72.92\%$ and
$77.08\%$ for the wider subsets. The corresponding average gaps over uniform
selection are $4.22\%$ and $7.48\%$, compared with $4.69\%$ and $8.59\%$.
These draws describe variation within the recorded banks, not independent
evidence from 1,000 new experiments. The wider subsets expand numerical spread
more clearly than oracle coverage. This analysis reuses archived branches and
does not collect a new 32 candidate bank.

\paragraph{Replay validation}
A follow-up control uses the frozen production source to repeat one candidate
continuation four times at the first \texttt{TurnOffStove} anchor. Its state
traces, continuation action hashes and terminal outcome agree across repeats.
The anchor images, candidate action and future hashes, continuation action
hashes and terminal outcome also match the archived candidate. Controller and
random number state digests differ across runs, so the historical reconstruction
check does not pass. A source audit identifies process dependent pointer bytes
in the controller digest and a truncated representation in the random number
state digest. The archive lacks the field values needed to determine whether
these encoding defects account for every difference. This control supports
repeatability of the tested continuation, not complete historical state
certification or validation of the successful alternatives.

\paragraph{Terminal outcome instrumentation}
A separate capture control executes two candidates twice at the same anchor.
Each branch runs its initial 32 actions and then the production continuation
for the remaining task budget. All four reach the 490 step budget from the
anchor without success, with 32 initial actions and 458 continuation steps.
Repeated captures, continuation action hashes and state traces agree within
each candidate. The initial 32 action execution differs from the 16 action
prefix used in the frozen bank, so these outcomes are not added to that bank.
This validates terminal label recording on a small control rather than
providing an outcome diverse comparison. A stop before a task terminal event
is recorded as an unknown outcome, not a failure.

\subsection{Full Chunk Selection Across Trajectory Phases}
\label[appendix]{app:full_chunk}

\paragraph{Collection protocol}
The fresh panel uses \texttt{CloseSingleDoor}, \texttt{OpenSingleDoor},
\texttt{PnPCounterToSink} and \texttt{TurnOffStove}, with episode indices 0 to
15 for each task. Scene seeds use the same deterministic rule as the earlier
bank, so the first three episode identities overlap. The trajectories and
candidate banks are newly collected, rather than disjoint held out scenes.
After collecting each behaviour trajectory, one eligible decision is selected
uniformly before candidate outcomes are collected. The final recorded decision
and decisions with at most 33 steps remaining are excluded. The resulting
64 anchors comprise 29 early, 21 middle and 14 late decisions. Each episode
contributes only one anchor, avoiding repeated anchors from one trajectory in
the uncertainty calculation.

The behaviour policy, initial candidate execution and subsequent continuation
all use 32 actions before replanning. With ten initial stabilisation steps,
the decision grid is $t=10+32d$. The candidate future image targets the state
after this initial chunk. Candidate seeds are fixed at 195 through 210. Branches
terminate at success or the task limit of 500 steps for the door and stove
tasks and 700 for the pick and place task. All 1,024 branches have observed
terminal outcomes, comprising 514 successes and 510 failures at the task
limit. There are no censored branches or recorded execution errors.

\paragraph{Scoring and outcome composition}
The score set and direction match the frozen comparison, using stored learned
value, current to future SSIM and the fixed Qwen3-VL image pair prompt.
Every scored row is joined by candidate identity and checked against its
recorded action and future hashes. The complete cache contains the first Yes
and No token logits in bfloat16 precision and their normalised probability.
All values are finite. The score is computed before terminal labels are joined,
and ties choose the lowest candidate index.

At $K=16$, 25 anchors contain only successes, 26 contain only failures, and 13
contain both. The mixed counts are four, three, four and two across the tasks
in the order listed above. As shown in \cref{tab:full_chunk_tasks}, this panel
contains useful choices in both door tasks as well as the manipulation tasks.
The two candidate counts in \cref{tab:full_chunk_selection} use nested prefixes
of the same banks, not separately sampled states or independent trials.

\begin{table*}[t]
\vspace*{5pt}
\centering
\caption{\textbf{Full chunk selection by task.}
  Every task contributes 16 decision states and 16 candidates per state.
  Entries are success percentages.}
\label{tab:full_chunk_tasks}
\footnotesize
\begin{tabular*}{\textwidth}{@{\extracolsep{\fill}}lrrrrr@{}}
\toprule
\textbf{Rule} & \textbf{Close door} & \textbf{Open door} & \textbf{Counter to sink} & \textbf{Turn off stove} & \textbf{Overall} \\
\midrule
Uniform & 91.80 & 66.41 & 34.77 & 7.81 & 50.20 \\
Learned value & 93.75 & 62.50 & 37.50 & 12.50 & 51.56 \\
SSIM & 93.75 & 62.50 & 37.50 & 6.25 & 50.00 \\
VLM progress & 93.75 & 68.75 & 31.25 & 6.25 & 50.00 \\
Outcome oracle & 100.00 & 75.00 & 43.75 & 18.75 & 59.38 \\
\bottomrule
\end{tabular*}
\end{table*}

\paragraph{Uncertainty and score ties}
We resample the 16 episodes within each of the four tasks, preserving all
candidate outcomes and scores at each anchor. We use 10,000 paired resamples
with seed 195 and report the central 95\% range. At $K=16$, the oracle gain
over uniform is $9.18\%$, with a range from $3.81\%$ to $15.23\%$. The
learned value gain is $1.37\%$, with a range from $-2.83\%$ to $5.76\%$.
The remaining oracle gap is positive for all three tested selectors under
this conditional resampling, as shown in \cref{tab:full_chunk_uncertainty}.
These ranges concern the recorded tasks and finite candidate banks.

VLM progress has 38 distinct recorded probabilities, with a tie for the top
score at 33 of 64 anchors. Averaging outcomes across the top tied candidates
gives $48.07\%$ expected success, compared with $50.00\%$ under the fixed
index rule. Learned value has one top tied anchor and changes from $51.56\%$
to $50.33\%$ under random ties. SSIM has no top ties and remains at $50.00\%$.
The small learned value advantage is therefore sensitive to tie handling.
For comparison, random tie resolution in the historical frozen VLM panel
changes success from $66.67\%$ to $67.08\%$, still below its uniform baseline.

\begin{table*}[t]
\vspace*{5pt}
\centering
\caption{\textbf{Uncertainty and tie sensitivity in full chunk selection.}
  Ranges are the central 95\% of 10,000 paired episode bootstrap resamples within each task.
  Changes and remaining oracle gaps are in percentage points.
  Random ties average over the top scoring tied candidates in each recorded bank.}
\label{tab:full_chunk_uncertainty}
\footnotesize
\begin{tabular*}{\textwidth}{@{\extracolsep{\fill}}clrrrr@{}}
\toprule
$K$ & \textbf{Rule} & \textbf{Change from uniform} & \textbf{Remaining oracle gap} & \textbf{Tied anchors} & \textbf{Random ties (\%)} \\
\midrule
4 & Learned value & -4.30 to +3.52 & 0.00 to 10.94 & 1 & 48.83 \\
 & SSIM & -4.30 to +3.52 & 0.00 to 10.94 & 0 & 50.00 \\
 & VLM progress & -3.12 to +2.34 & 0.00 to 10.94 & 23 & 50.00 \\
\midrule
16 & Learned value & -2.83 to +5.76 & 1.56 to 15.62 & 1 & 50.33 \\
 & SSIM & -3.81 to +3.32 & 3.12 to 17.19 & 0 & 50.00 \\
 & VLM progress & -2.73 to +2.15 & 3.12 to 17.19 & 33 & 48.07 \\
\bottomrule
\end{tabular*}
\end{table*}

\paragraph{Validation and limits}
A fresh control repeats candidate zero four times from one
\texttt{CloseSingleDoor} anchor. Its continuation actions, state traces and
terminal success at step 189 agree within the process. The same candidate's
continuation action sequence and outcome also match between that control and
the collected block. This is evidence for the tested continuation, not
certification of every state across processes. The saved package does not
contain full serialised simulator snapshots or the realised images after the
initial 32 actions. No matched predicted versus realised image experiment was
completed, and old terminal labels are not reused for reconstructed states.
Since both the decision sampling and execution horizon differ from the frozen
prefix panel, differences between the panels cannot be attributed to either
change alone.

\subsection{Additional Selector Results and Input Diagnostics}
\label[appendix]{app:complete_selector_results}

\subsubsection{Task Difficulty Groups}
\label{app:difficulty_groups}
We define task difficulty using only the task level success of the separately
run $K=1$ fixed candidate reference. Hard tasks have success below $50\%$,
medium tasks have success from $50\%$ to below $80\%$, and easy tasks have
success of at least $80\%$. This fixed rule yields 6 hard, 11 medium, and 7 easy
tasks. The assignment is held constant for every selection rule in
\cref{tab:closed_loop_by_difficulty} and does not depend on selector
performance.

\paragraph{Historical feature probes}
In \cref{tab:feature_probes_icra}, we preserve the results of an earlier linear
probe on 2,400 demonstration anchors, with four candidates per anchor. Its target is the
candidate with the lowest mean absolute action error to the demonstration
chunk. No candidate is executed to produce this label. Five cross validation
folds keep candidates from one anchor together, but do not constitute a held
out trajectory test. The full fit has AUC $0.499$ and top 1 agreement $24.3\%$.

The implementation audit identifies seven features that average the current
observation with predicted images, five temporal or mixed reductions, and two
remaining inputs, learned value and final-frame SSIM. The original clip
manifest is unavailable, leaving its exact frame count unresolved. Aesthetic
quality, CLIP physics and CLIP task alignment in \cref{fig:candidate_spread}
therefore remain labelled as recorded input statistics rather than clean
prediction-only measurements.

A CPU refit reproduces the full AUC of $0.499476$. Restricting the input to
learned value and SSIM produces zero coefficients in all five folds. Its AUC
of $0.500$ and top 1 agreement of $25.5\%$ follow from constant predictions
and choosing candidate zero. Reducing the optimiser tolerance from $10^{-4}$
to $10^{-10}$ yields nonzero coefficients and AUC $0.499964$. A separate split
that keeps each of the 240 source trajectories together gives full-feature
AUC $0.500207$. These checks concern demonstration-action matching, not
recognition of successful branches. The pixel change and value-only family
fits also have zero coefficients. The table retains those fits with explicit
markers rather than treating them as independent negative results.

\begin{table}[t]
\vspace*{5pt}
\centering
\caption{\textbf{Historical demonstration-action matching probes.}
  The target is the candidate closest to the demonstration action chunk, not
  task success. Top 1 chance is $25\%$ and chance AUC is $0.50$. These archived
  fits include affected or unsupported inputs and are not a corrected metric
  comparison.}
\label{tab:feature_probes_icra}
\scriptsize
\setlength{\tabcolsep}{2.0pt}
\renewcommand{\arraystretch}{0.94}
\begin{tabular*}{\columnwidth}{@{\extracolsep{\fill}}lrrr@{}}
\toprule
\textbf{Probe input} & \textbf{AUC} & \textbf{Top 1} & \textbf{Interval} \\
\midrule
\multicolumn{4}{@{}l}{\itshape Feature family}\\[-1pt]
\rowcolor{tablegray}[0pt][0pt]All features & .499 & 24.3 & $22.6$--$25.9$ \\
Quality + dynamics & .499 & 24.1 & $22.4$--$25.8$ \\
Task + physics & .500 & 24.5 & $22.8$--$26.3$ \\
Pixel change$^\dagger$ & .500 & 25.5 & $23.8$--$27.3$ \\
Learned value$^\dagger$ & .500 & 25.5 & $23.8$--$27.3$ \\
\addlinespace[1pt]
\multicolumn{4}{@{}l}{\itshape Decisions with larger action differences}\\[-1pt]
Top 50\% & .498 & 23.2 & $20.8$--$25.6$ \\
Top 25\% & .499 & 25.2 & $21.8$--$28.5$ \\
Top 10\% & .503 & 29.6 & $23.8$--$35.4$ \\
\addlinespace[1pt]
\multicolumn{4}{@{}l}{\itshape Additional refit checks}\\[-1pt]
Value + SSIM$^\dagger$ & .500 & 25.5 & $23.8$--$27.3$ \\
All features, trajectory split & .500 & 23.2 & $21.4$--$24.9$ \\
\bottomrule
\end{tabular*}
\vspace{2pt}
\parbox{\columnwidth}{\scriptsize
$^\dagger$ Zero fitted coefficients and constant predictions. Top 1 intervals
retain the original anchor resampling, including for the trajectory-split
sensitivity row.}
\end{table}

\paragraph{Closed loop selectors}
The results in \cref{tab:closed_loop_full_icra} use the released
Cosmos-Policy-RoboCasa-Predict2-2B checkpoint. We retain the evaluator set in
the main comparison and report constant repeated-frame controls separately
within the table. Additional reward-model runs remain in the experiment
archive but are not part of this comparison.
The five reported slot input selectors contribute $6{,}000$ episodes from
24 tasks with 50 trials each, using $K=4$, seed 195.
All five success totals match their 120 saved run summaries per
selector. Manifests record five input images, comprising the current observation
and four images decoded from one future camera slot. The slot images represent one future
timestep, not a temporal sequence. Intervals use $10{,}000$ paired task resamples against the
fixed candidate reference, without adjustment across selectors.

No reported selector in this slot input panel improves the reference point
estimate, and all five intervals include zero. These results describe the
evaluated input adaptation, not scoring on a predicted temporal rollout.

The log census for these five selectors covers 139,893 decisions and 559,572 scores. All
recorded scores are finite, and every selected index matches the first
maximum of its score vector. VLM controllability ties all candidates at
$60.4\%$ of decisions and selects candidate zero at $80.0\%$.
VLM progress value ties at $79.1\%$, returns all zeros at $40.5\%$, and selects
candidate zero at $89.1\%$. It reads only the last input image. None of the five rules is constant
at every decision.

The summaries contain no scorer-failure count, and raw VLM responses were not
saved. Two parsers can return zero on a failed parse, so parse failures cannot
be distinguished from genuine zero responses. The finite-score check does not
resolve that missing information. The legacy panel
retains verified task aggregates but lacks the original per decision records.
In particular, the historical direction of foreground centroid jerk cannot be
confirmed from the later runtime source.

\begin{table*}[t]
\vspace*{5pt}
\centering
\caption{\textbf{Extended closed loop selector results.}
  Every selector is evaluated on 24 tasks with 50 trials per task, except the
  learned value, which pools two independently run panels. Changes and
  task resampled intervals use the separately run fixed candidate reference at $65.9\%$.
  A dagger marks a constant rule, a double dagger marks majority ties in the
  complete 24-task log audit, and a
  section sign marks an unverified historical score direction.}
\label{tab:closed_loop_full_icra}
\footnotesize
\setlength{\tabcolsep}{4.2pt}
\renewcommand{\arraystretch}{0.96}
\begin{tabular*}{\textwidth}{@{\extracolsep{\fill}}p{0.35\textwidth}rrrr@{}}
\toprule
\textbf{Selection rule} & \textbf{Episodes} & \textbf{Success} &
\textbf{Change} & \textbf{Task resampled interval} \\
\midrule
Fixed candidate $K\!=\!1$ & 1,200 & 65.9 & reference & n.a. \\
\midrule
\multicolumn{5}{@{}l}{\itshape Repeated frame and learned value panel}\\
\rowcolor{tablegray}[0pt][0pt]Learned value & 2,400 & 66.8 & $+0.9$ & $-1.5\ \text{to}\ +3.2$ \\
Background consistency & 1,200 & 66.7 & $+0.8$ & $-1.6\ \text{to}\ +3.2$ \\
Laplacian sharpness & 1,200 & 66.1 & $+0.2$ & $-2.0\ \text{to}\ +2.3$ \\
Valid colour range$^\dagger$ & 1,200 & 66.0 & $+0.1$ & $-0.6\ \text{to}\ +0.8$ \\
Peak to mean motion$^\dagger$ & 1,200 & 66.0 & $+0.1$ & $-0.3\ \text{to}\ +0.5$ \\
Gradient sharpness & 1,200 & 65.8 & $-0.1$ & $-2.3\ \text{to}\ +2.0$ \\
Spatial complexity & 1,200 & 65.8 & $-0.1$ & $-1.8\ \text{to}\ +1.8$ \\
Motion region overlap$^\dagger$ & 1,200 & 65.8 & $-0.2$ & $-0.7\ \text{to}\ +0.3$ \\
Aesthetic quality & 1,200 & 65.7 & $-0.3$ & $-2.3\ \text{to}\ +1.6$ \\
SSIM & 1,200 & 65.5 & $-0.4$ & $-2.8\ \text{to}\ +1.8$ \\
Subject consistency & 1,200 & 65.3 & $-0.6$ & $-2.8\ \text{to}\ +1.5$ \\
Image quality & 1,200 & 65.3 & $-0.6$ & $-2.7\ \text{to}\ +1.3$ \\
Frame difference & 1,200 & 65.3 & $-0.7$ & $-2.9\ \text{to}\ +1.3$ \\
Foreground centroid jerk$^\S$ & 1,200 & 65.1 & $-0.8$ & $-2.8\ \text{to}\ +1.4$ \\
Temporal flicker & 1,200 & 64.7 & $-1.3$ & $-3.7\ \text{to}\ +1.0$ \\
\midrule
\multicolumn{5}{@{}l}{\itshape Decoded camera slot input}\\
VLM controllability$^\ddagger$ & 1,200 & 65.6 & $-0.3$ & $-3.3\ \text{to}\ +2.8$ \\
VLM progress value$^\ddagger$ & 1,200 & 65.6 & $-0.3$ & $-3.0\ \text{to}\ +2.4$ \\
VQAScore & 1,200 & 64.9 & $-1.0$ & $-4.2\ \text{to}\ +2.3$ \\
VLM progress & 1,200 & 64.7 & $-1.3$ & $-4.7\ \text{to}\ +2.3$ \\
VideoPhy & 1,200 & 64.3 & $-1.7$ & $-4.8\ \text{to}\ +1.3$ \\
\bottomrule
\end{tabular*}
\end{table*}

\paragraph{Metrics that cannot rank the repeated frame input}
Three temporal rules are algebraically constant when a decoded frame is tiled
over time, and valid colour range is constant after normalisation regardless
of tiling. Three of these rules were included in the closed loop sweep and are
marked with daggers in \cref{tab:closed_loop_full_icra}. Motion smoothness was
not run online. Their numerical outcomes cannot be attributed to their scores
because all candidates tie. These were independent end to end runs whose
candidate tensors were not matched by hash across conditions, so their small
differences are not a same bank random or fixed index control.

\begin{table*}[t]
\vspace*{5pt}
\centering
\caption{\textbf{Supporting diagnostics.}
  Panel (a) reports repeated frame rules that tie within every decision.
  \emph{(b)} Historical outcomes from the weaker snapshot procedure described
  in \cref{app:outcome_detail}. A mixed state has both successful and failed branches.}
\label{tab:diagnostic_support_icra}
\scriptsize
\begin{minipage}[t]{0.48\textwidth}
\centering
\textit{(a) Degenerate repeated frame metrics}\par\vspace{2pt}
\renewcommand{\arraystretch}{0.96}
\begin{tabular*}{\linewidth}{@{\extracolsep{\fill}}lrl@{}}
\toprule
\textbf{Metric} & \textbf{Value} & \textbf{Reason} \\
\midrule
Motion region overlap & 0.000 & Zero motion map \\
Peak to mean motion & 8.000 & $\max/\mathrm{mean}=T-1$ \\
Motion smoothness & 0.429 & Identity $1-4/7$ \\
Valid colour range & 1.000 & Frames clipped to $[0,1]$ \\
\bottomrule
\end{tabular*}
\end{minipage}
\hfill
\begin{minipage}[t]{0.48\textwidth}
\centering
\textit{(b) Paired outcomes by task}\par\vspace{2pt}
\setlength{\tabcolsep}{3pt}
\renewcommand{\arraystretch}{0.94}
\begin{tabular*}{\linewidth}{@{\extracolsep{\fill}}lrrr@{}}
\toprule
\textbf{Task} & \textbf{States} & \textbf{Mixed} & \textbf{Rate} \\
\midrule
CloseDrawer & 124 & 0 & 0.0 \\
OpenDrawer & 70 & 2 & 2.9 \\
PnPCounterToCab & 53 & 3 & 5.7 \\
CoffeePressButton & 10 & 0 & 0.0 \\
TurnOnSinkFaucet & 76 & 1 & 1.3 \\
PnPSinkToCounter & 51 & 5 & 9.8 \\
\midrule
\textbf{All tasks} & \textbf{384} & \textbf{11} & \textbf{2.9} \\
\bottomrule
\end{tabular*}
\end{minipage}
\end{table*}

\paragraph{Complete decoded slot diagnostic}
Decoding Cosmos Policy produces 41 pixel frames spanning the latent slots for
proprioception, cameras, actions, and value. The primary camera future occupies
frames 29 to 32. During training, this slot contains four copies of one camera
image at one timestep~\citep{cosmospolicy2026}. The four decoded images are
therefore not four successive predicted timesteps. The reference-image
correction supplies the real current observation where a score requires it.
For model-free absolute image properties, the revised reduction excludes that
observation. The audit subsequently found that aesthetic quality, CLIP physics
and CLIP task alignment still average it with the predicted frames. Their
recorded statistics are retained with this distinction. No reduction creates a
temporal rollout from these slot images.

In \cref{tab:corrected_proxy_audit}, we report the audited export for 194
decisions in 12 task and trial groups. The target is mean pixel error to one observation
after the first action of candidate zero, shared by all candidates. Mean error
is $0.087922$ under uniform selection, $0.087555$ for the lowest error candidate,
and $0.087700$ under SSIM. Thus, the available reduction is $0.42\%$ of uniform
error and SSIM achieves a $0.25\%$ reduction, recovering $60.5\%$ of that gap.
Mean SSIM recovers $78.8\%$. These reference-image scores favour similarity to
the current image, not departure from it or task progress.

Both decision and whole-group resampling use 2,000 draws and preserve the same
point estimate. SSIM's recovery range is $45.3$ to $73.5$ with decision
resampling and $34.6$ to $73.3$ with group resampling. CLIP task alignment has
negative recovery under both schemes for its recorded reduction. A shared
current-frame term in a linear per-frame mean changes score values and spread,
but preserves within-decision ordering when the future frames, weights and
per-frame scores are otherwise unchanged. We test this directly by comparing
the four predicted slot images alone with the current observation followed by
those same images. Aesthetic quality, CLIP physics and CLIP task alignment
preserve the selected candidate and full ordering at all 194 decisions.
Including the current observation reduces their within decision standard
deviation to $0.8$ of the prediction only value, as expected for a mean over
five rather than four images. Thus, the shared observation changes spread
without changing the measured ranking or pixel error recovery. The comparison
does not cover the separate 2,400-anchor probe panel, and the historical arm
does not retain weight or source metadata. Results are exploratory and unadjusted
across scores. Unsupported temporal and mixed reductions remain labelled as
such even when their numerical recovery is high. Corrected VideoPhy scores
are unavailable in this export, so its historical offline values are not
substituted. The online results are separate runs.

\subsection{Transfer Across Models and Benchmarks}
\label[appendix]{app:transfer}

Historical score-transfer calculations between DreamZero with
DROID~\citep{khazatsky2024droid} and Cosmos Policy with RoboCasa are excluded
from the active comparison while their input and target dependencies are
unresolved. The action statistics below are computed from the sampled actions
and do not use those visual-score recoveries.

In \cref{fig:transfer}, the shared diversity audit covers Cosmos Policy on
RoboCasa, LingBot-VA on LIBERO-Long, and RynnVLA on LIBERO-10~\citep{liu2023libero}. Their
within decision to marginal action spread ratios are $0.016$, $0.049$, and
$0.093$, respectively. Each ratio uses the same dimensionless standard
deviation calculation, although the underlying action parameterisations and
benchmarks differ.

The secondary closed loop checks have different limitations. Learned value
selection with Cosmos Policy changes macro success on four LIBERO suites from $98.05\%$ to
$98.00\%$. RynnVLA's valid $K\!=\!1$ and random-$K\!=\!4$ conditions reach
$85.6\%$ and $85.0\%$. Its attempted imagined-frame scorer is excluded because
all candidate scores failed and the implementation always selected candidate
zero. LingBot-VA reaches $97.0\%$ at $K\!=\!1$ and $97.5\%$ under both random
selection and a task progress VLM selector. The latter contains 172 episodes.
None of the task resampled intervals exclude zero. These settings therefore
provide compatibility aware breadth, not matched replications of the primary
Cosmos Policy experiment.

\subsection{Sampling and Evaluation Scale}
\label{app:robustness}

\begin{table}[t]
\vspace*{5pt}
\centering
\caption{\textbf{Sampler intervention at $K\!=\!4$.}
  The intervention changes only the Cosmos sampler. Diversity is within
  decision action variance divided by marginal action
  variance.}
\label{tab:robustness_icra}
\scriptsize
\setlength{\tabcolsep}{1.8pt}
\renewcommand{\arraystretch}{0.96}
\begin{tabular*}{\columnwidth}{@{\extracolsep{\fill}}lrr@{}}
\toprule
\textbf{Sampler} & \textbf{$n$} & \textbf{Action variance ratio} \\
\midrule
5 steps & 190 & 0.0043 \\
1 step & 188 & 0.0001 \\
5 steps + variance & 190 & 0.0045 \\
1 step + variance & 187 & 0.0001 \\
\bottomrule
\end{tabular*}
\end{table}

\paragraph{Sampler intervention}
We next change only the sampler while holding Cosmos Policy, the benchmark,
$K\!=\!4$, and the score panel fixed. Action diversity is the mean
within decision action variance divided by marginal action variance, averaged
over action dimensions. As shown in \cref{tab:robustness_icra}, this ratio
remains below $0.005$ across four regimes and approximately 190 decisions
per regime. These changes do not substantially expand the
numerical action spread in the recorded regimes. Scorer-dependent recovery
values are omitted because their input dependencies remain unresolved. This intervention does not
establish that increasing diversity would improve control.
The recovered configurations use either one or five denoising steps, with
variance scaling disabled or enabled, as labelled in the table. All four
use the same checkpoint snapshot and candidate seed schedule. At environment
step $t$, candidate $i$ uses seed $195+t+i$, and the variance draw uses that
same seed when enabled. The checkpoint revision is recorded, but the historical
runtime source was not pinned. The configuration recovery confirms the
displayed sampler settings without restoring the omitted visual score claims.

\paragraph{Evaluation scale}
Among 500 subsamples, a screen with four tasks and ten trials recovers the leading
selector from the complete evaluation only $10\%$ of the time and has mean Kendall
$\tau=0.068$. Using all 24 tasks with ten trials gives $13\%$ recovery and
$\tau=0.141$. As shown in \cref{fig:evaluation_scale}, rankings stabilise only
near the complete $24\times50$ grid.

\begin{figure}[!htbp]
\centering
\includegraphics[width=0.88\columnwidth]{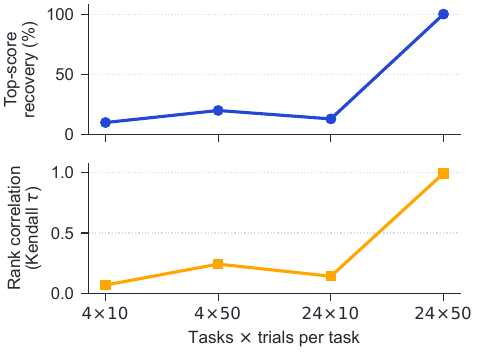}
\caption{\textbf{Selector-ranking stability as evaluation grows.}
  Small screens rarely recover the leading selector from the complete evaluation and
  have low rank correlation with the complete evaluation.}
\label{fig:evaluation_scale}
\end{figure}

\begin{table*}[!t]
\vspace*{5pt}
\centering
\caption{\textbf{Audited recorded-observation diagnostic.}
  Recovery is the fraction of the uniform-to-lowest-pixel-error gap recovered
  by a score, not task success. All available rows use 194 decisions from
  12 task and trial groups. Both ranges use 2,000 resamples.}
\label{tab:corrected_proxy_audit}
\footnotesize
\setlength{\tabcolsep}{3pt}
\renewcommand{\arraystretch}{0.94}
\begin{tabular*}{\textwidth}{@{\extracolsep{\fill}}p{0.32\textwidth}crrr@{}}
\toprule
\textbf{Score} & \textbf{Status} & \textbf{Recovery (\%)} &
\textbf{Decision range} & \textbf{Group range} \\
\midrule
\multicolumn{5}{@{}l}{\itshape Corrected endpoint and reference-image computations}\\
Final frame entropy & C & $-15.4$ & $-32.7\ \text{to}\ +3.9$ & $-38.7\ \text{to}\ -0.7$ \\
Final gradient sharpness & C & $-11.0$ & $-28.9\ \text{to}\ +6.0$ & $-30.3\ \text{to}\ -0.5$ \\
Final Laplacian sharpness & C & $-28.4$ & $-44.8\ \text{to}\ -12.0$ & $-38.4\ \text{to}\ -9.2$ \\
Final observation histogram correlation & C & $+11.3$ & $-10.5\ \text{to}\ +31.3$ & $+0.1\ \text{to}\ +28.0$ \\
Final start edge correlation & C & $+34.7$ & $+17.5\ \text{to}\ +52.4$ & $+8.8\ \text{to}\ +49.5$ \\
Gradient sharpness & C & $-25.3$ & $-41.2\ \text{to}\ -8.7$ & $-40.7\ \text{to}\ -12.3$ \\
Laplacian sharpness & C & $-31.9$ & $-49.1\ \text{to}\ -13.6$ & $-49.8\ \text{to}\ +1.8$ \\
Mean frame entropy & C & $-29.1$ & $-44.8\ \text{to}\ -13.1$ & $-47.7\ \text{to}\ -14.4$ \\
Mean observation histogram correlation & C & $+12.9$ & $-8.9\ \text{to}\ +33.3$ & $+1.2\ \text{to}\ +34.4$ \\
Mean SSIM & C & $+78.8$ & $+69.4\ \text{to}\ +87.2$ & $+63.8\ \text{to}\ +85.3$ \\
Mean start edge correlation & C & $+45.5$ & $+29.8\ \text{to}\ +61.3$ & $+18.3\ \text{to}\ +62.6$ \\
Spatial complexity & C & $-14.4$ & $-35.6\ \text{to}\ +5.0$ & $-34.9\ \text{to}\ -2.3$ \\
SSIM & C & $+60.5$ & $+45.3\ \text{to}\ +73.5$ & $+34.6\ \text{to}\ +73.3$ \\
Uncertainty component & C & $-0.8$ & $-24.7\ \text{to}\ +23.7$ & $-35.4\ \text{to}\ +20.6$ \\
VLM controllability & C & $-17.5$ & $-35.0\ \text{to}\ +0.5$ & $-28.5\ \text{to}\ +4.7$ \\
VLM progress & C & $-6.1$ & $-26.2\ \text{to}\ +14.1$ & $-21.2\ \text{to}\ +19.4$ \\
VLM progress value & C & $-7.0$ & $-26.8\ \text{to}\ +15.4$ & $-16.7\ \text{to}\ +13.7$ \\
VQAScore & C & $+5.5$ & $-10.9\ \text{to}\ +22.5$ & $-13.3\ \text{to}\ +17.3$ \\
\midrule
\multicolumn{5}{@{}l}{\itshape Unsupported temporal, mixed or slot-dispersion measurements}\\
Background consistency & U & $+1.0$ & $-16.1\ \text{to}\ +18.5$ & $-22.7\ \text{to}\ +19.2$ \\
Dynamic degree & U & $-85.5$ & $-95.9\ \text{to}\ -74.9$ & $-101.8\ \text{to}\ -75.1$ \\
Edge flicker & U & $+46.5$ & $+30.4\ \text{to}\ +61.1$ & $+29.3\ \text{to}\ +57.9$ \\
Foreground area variation & U & $+6.1$ & $-11.2\ \text{to}\ +23.7$ & $-17.4\ \text{to}\ +18.1$ \\
Foreground centroid jerk & U & $-0.9$ & $-19.2\ \text{to}\ +17.7$ & $-16.6\ \text{to}\ +6.7$ \\
Foreground spread variation & U & $+17.2$ & $-2.0\ \text{to}\ +33.9$ & $-18.6\ \text{to}\ +41.6$ \\
Frame difference & U & $+87.5$ & $+80.9\ \text{to}\ +92.6$ & $+76.3\ \text{to}\ +93.0$ \\
Motion centroid jerk & U & $+2.7$ & $-17.0\ \text{to}\ +22.2$ & $-8.9\ \text{to}\ +23.1$ \\
Motion compactness & U & $-12.6$ & $-28.5\ \text{to}\ +3.5$ & $-35.7\ \text{to}\ +0.8$ \\
Motion focus & U & $-4.1$ & $-24.8\ \text{to}\ +18.3$ & $-23.2\ \text{to}\ +25.7$ \\
Motion magnitude & U & $-87.6$ & $-97.9\ \text{to}\ -77.1$ & $-103.4\ \text{to}\ -77.1$ \\
Motion region overlap & U & $+9.7$ & $-11.3\ \text{to}\ +29.7$ & $-11.3\ \text{to}\ +43.5$ \\
Motion smoothness & U & $-11.1$ & $-28.3\ \text{to}\ +5.3$ & $-28.6\ \text{to}\ +0.4$ \\
Peak to mean motion & U & $-31.5$ & $-47.8\ \text{to}\ -13.0$ & $-49.6\ \text{to}\ +3.4$ \\
Predicted pixel variance & U & $-89.5$ & $-97.9\ \text{to}\ -81.2$ & $-100.6\ \text{to}\ -83.3$ \\
Prediction history consistency & U & $+1.9$ & $-18.1\ \text{to}\ +23.1$ & $-14.8\ \text{to}\ +30.4$ \\
Semantic component & U & $+32.7$ & $+12.6\ \text{to}\ +50.9$ & $+9.9\ \text{to}\ +43.5$ \\
Subject consistency & U & $+32.7$ & $+12.6\ \text{to}\ +50.9$ & $+9.9\ \text{to}\ +43.5$ \\
Temporal acceleration & U & $+74.1$ & $+61.9\ \text{to}\ +83.5$ & $+43.3\ \text{to}\ +89.9$ \\
Temporal flicker & U & $+95.7$ & $+92.9\ \text{to}\ +97.8$ & $+94.7\ \text{to}\ +96.8$ \\
\midrule
\multicolumn{5}{@{}l}{\itshape Diagnostic or provisional computations}\\
Candidate diversity & D & $-15.2$ & $-33.2\ \text{to}\ +6.6$ & $-27.3\ \text{to}\ +8.8$ \\
Colour overflow & D & $-15.2$ & $-33.2\ \text{to}\ +6.6$ & $-27.3\ \text{to}\ +8.8$ \\
Combined semantic uncertainty & D & $+26.2$ & $+4.6\ \text{to}\ +45.6$ & $-8.6\ \text{to}\ +45.4$ \\
Valid colour range & D & $-15.2$ & $-33.2\ \text{to}\ +6.6$ & $-27.3\ \text{to}\ +8.8$ \\
\midrule
\multicolumn{5}{@{}l}{\itshape Frame means with verified ranking agreement}\\
Aesthetic quality & R & $+18.3$ & $-2.4\ \text{to}\ +37.6$ & $-0.7\ \text{to}\ +28.6$ \\
CLIP physics & R & $+15.3$ & $-4.9\ \text{to}\ +34.2$ & $+3.4\ \text{to}\ +21.9$ \\
CLIP task alignment & R & $-30.8$ & $-46.5\ \text{to}\ -15.2$ & $-43.6\ \text{to}\ -7.0$ \\
\midrule
\multicolumn{5}{@{}l}{\itshape Corrected export unavailable}\\
VideoPhy & N & n.a. & n.a. & n.a. \\
\bottomrule
\end{tabular*}
\vspace{2pt}
\parbox{\textwidth}{\scriptsize
C denotes a corrected or verified unchanged computation, not validation of
task relevance. U denotes an unsupported interpretation on the decoded-slot
interface. D denotes diagnostic or provisional status. R retains a recorded
current-plus-prediction mean whose ranking agrees with the prediction-only mean
at all 194 decisions. N denotes a missing corrected export. High recovery
does not validate U or D rows as temporal, physical or uncertainty measurements.}
\end{table*}

\subsection{Counterfactual Outcomes and Retrospective Selection}
\label{app:outcome_detail}

\paragraph{Counterfactual outcome protocol}
This historical panel uses a weaker snapshot procedure than the frozen
$K=16$ bank. It saves the MuJoCo simulator state~\citep{todorov2012mujoco}
and four environment scalars, but not the full controller, observable and
random number state. For each candidate, it restores those saved quantities,
executes the 16 step action chunk, and resumes greedy Cosmos Policy. Six tasks
use seed 195 and contribute 384 decision states, or 1,536 branches. No recorded
restoration or identical-candidate control accompanies this panel. Its outcomes
therefore describe that procedure rather than establish an isolated effect of
candidate choice.

All four branches succeed at $59.6\%$ of states and all fail at $37.5\%$.
The remaining 11 states are outcome diverse. As shown in panel (b) of
\cref{tab:diagnostic_support_icra}, they concentrate on the two pick and place
tasks, while \texttt{CloseDrawer} and \texttt{CoffeePressButton} have none.

An incomplete episode level oracle pilot finished ten
\texttt{CoffeePressButton} and two \texttt{CloseDrawer} episodes before the
remaining runs exceeded the compute budget. These unpaired counts do not
estimate selection headroom. Likewise, nested decision states do not turn the
$2.9\%$ mixed state frequency into an episode level upper bound.

\begin{samepage}
\paragraph{Retrospective selector analyses}
The following values are retained from the historical exploratory report.
The complete permutation and held out routing procedure has not been
reconstructed from the deposited package, so these values are not used to
support the main conclusions. Nine tasks show no taskwise selection gain.
The largest reported increase is $18\%$ on
\texttt{TurnOffSinkFaucet} under motion centroid jerk, but the rule harms five
other tasks. Taskwise selection raises macro success by $3.9\%$. Permuting task
and selector identities yields $8.9\%$, while leave one task out selection
reaches $0.75\%$.
\end{samepage}

\fi

\end{document}